%% file: main.tex
\documentclass[10pt,twocolumn,letterpaper]{article}

\usepackage{iccv}
\usepackage{times}
\usepackage{epsfig}
\usepackage{graphicx}
\usepackage{amsmath}
\usepackage{amssymb}
\usepackage{multirow}
\usepackage{soul}
\usepackage{color}
\usepackage{adjustbox}
\usepackage{caption}
\usepackage[table]{xcolor}
\definecolor{darkgreen}{rgb}{0,0.45,0}
\definecolor{darkred}{rgb}{0.45,0,0}

\DeclareMathOperator*{\argmax}{arg\,max}

\usepackage[breaklinks=true,bookmarks=false]{hyperref}

\iccvfinalcopy 

\def\iccvPaperID{9498} 

\ificcvfinal\fi

\begin{document}

\title{Predicting with Confidence on Unseen Distributions}

\author{Devin Guillory\\
UC Berkeley\\
{\tt\small dguillory@berkeley.edu}
\and
Vaishaal Shankar\\
Amazon\\
{\tt\small vaishaal@amazon.com}

\and
Sayna Ebrahimi\\
UC Berkeley\\
{\tt\small sayna@berkeley.edu}
\and
Trevor Darrell \thanks{equal contribution}\\
UC Berkeley\\
{\tt\small trevor@eecs.berkeley.edu}
\and

Ludwig Schmidt \footnotemark[1]\\
Toyota Research Institute\\
{\tt\small ludwigschmidt2@gmail.com}
}

\maketitle
\ificcvfinal\thispagestyle{empty}\fi

\begin{abstract}
Recent work has shown that the accuracy of machine learning models can vary substantially when evaluated on a distribution that even slightly differs from that of the training data.
As a result, predicting model performance on previously unseen distributions without access to labeled data is an important challenge with implications for increasing the reliability of machine learning models. 
In the context of distribution shift, distance measures are often used to adapt models and improve their performance on new domains, however accuracy estimation is seldom explored in these investigations.
Our investigation determines that common distributional distances such as Frechet distance or Maximum Mean Discrepancy, fail to induce reliable estimates of performance under distribution shift.
On the other hand, we find that our proposed difference of confidences ($DoC$) approach  yields successful estimates of a classifier's performance over a variety of shifts and model architectures. 
Despite its simplicity, we observe that $DoC$ outperforms other methods across synthetic, natural, and adversarial distribution shifts, reducing error by ($>46\%$) on several realistic and challenging datasets such as ImageNet-Vid-Robust and ImageNet-Rendition.

\end{abstract}

\section{Introduction}

\begin{figure}[t]
    \centering
    \includegraphics[width=0.5\textwidth]{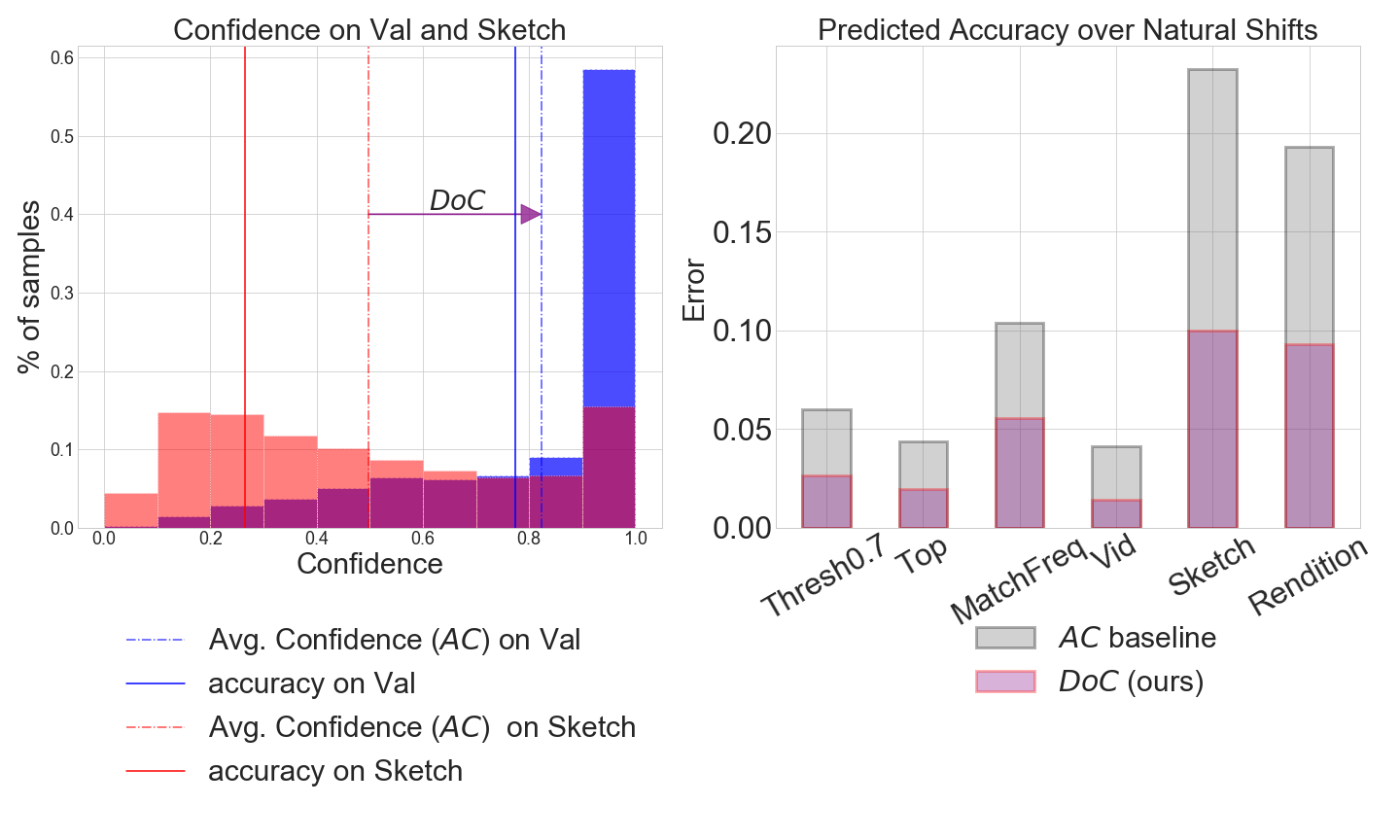}
    \caption{
    The average confidence ($AC$) over a distribution is a natural predictor of a model's accuracy. Models are often poorly calibrated, and while $AC$ provides an overly optimistic prediction of accuracy, the difference of confidences ($DoC$) between distributions provides a useful estimate of accuracy changes. 
    (\textbf{Left}) Confidence histograms of a ResNet-101 over the ImageNet validation set and ImageNet-Sketch. Average confidences ($AC$) are higher than the corresponding accuracies. Our method, difference of confidences ($DoC$) can be computed from $AC$ over the validation set and ImageNet-Sketch. 
    (\textbf{Right}) shows the average confidences ($AC$) as predictors of accuracy over various natural distribution shifts, contrasted with difference of confidences ($DoC$). $DoC$ consistently improves over the baseline. 
    }
    \label{fig:main_figure}
\end{figure}

Even under the best of conditions, machine learning models are still susceptible to large variations in performance whenever the test data does not come from the same distribution as the training data. 
For instance, recent dataset replication studies have shown that despite concerted efforts to closely mirror the data generating process of the training set, model accuracy still changed substantially on the new test sets \cite{recht2019imagenet, yadav2019cold, miller2020effect}. 
Machine learning models deployed in real-world environments invariably encounter different data distributions.
Hence reliable estimates of how well a model will perform on a new test set are critical. 
For instance, if a practitioner wants to use an X-ray classifier on data from a new hospital, they require a good estimate of their classifier’s performance on this previously unseen distribution to have confidence in the results. 
In some settings it maybe prohibitively costly to acquire new labeled data each time a model encounters a distribution shift.
In order to mitigate the consequences of unexpected performance changes, we explore how to best predict changes in a model’s accuracy when we only have access to unlabeled data. 

\par
We investigate the underexplored problem of Automatic model Evaluation \cite{deng2021labels} over various model architectures and forms of natural and synthetic distribution shift.
Distributional distances such as Frechet distance \cite{heusel2017gans, zilly2019frechet, deng2021labels}, Maximum Mean Discrepancy (MMD) \cite{borgwardt2006integrating} , and discriminative discrepancy \cite{ben2006analysis, ganin2016domain, glorot2011domain}, which are common tools for alignment in domain adaptation, are evaluated as features in a regression model for predicting accuracy on unseen distributions.
Recognizing the relationship of this problem with that of predictive uncertainty, we also evaluate the utility of model confidences and entropy at predicting accuracy.
\par
Learning our regression models over a set of synthetic distribution shifts, we show that common distributional distances fail to reliably predict accuracy changes on natural distribution shifts.
While most approaches are able to encode useful information about held-out forms of synthetic distribution shift, they do not produce useful encodings for natural distribution shifts, where a regression-free average confidence baseline $AC$ outperforms them.

\begin{figure}[t]
    \centering
    \includegraphics[width=\linewidth]{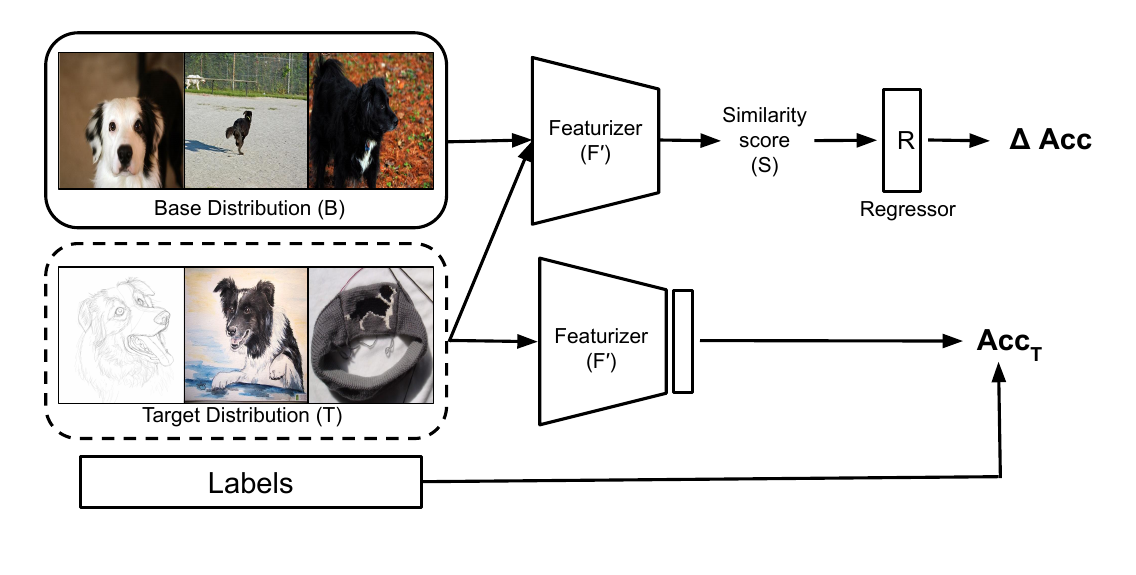}
    \caption{
    \noindent \textbf{Illustration of our proposed method.}
    In order to estimate a model $F$'s accuracy on a previously unseen and unlabeled target distribution $T$, we feed examples from the base distribution $B$ (usually the training distribution) and the target distribution $T$ into a featurizer $F^\prime$.
    In the calibration stage, we first compute a distributional difference, $S$, from the outputs of $F^\prime$ on various calibration sets and learn a regressor $R$ which predicts the difference in model accuracies from the distributional difference, i.e.,  $\Delta\text{Acc}=R(S)$.
    When we later encounter a new unlabeled target set, we again compute the distributional difference and then use the learned regressor to estimate the accuracy difference.
    Adding the estimated accuracy difference to the accuracy on the base distribution $B$ then yields the estimated accuracy on the target distribution $T$.
    This setup enables us to both accurately predict performance over unseen distribution shifts, and to better understand the quality of various distributional differences.}
    \label{fig:sec_figure}
\end{figure}

\par
Surprisingly, however, we discover that  a substantial amount of information about both synthetic and natural distribution shifts is encoded in the difference of confidences ($DoC$) of the classifier's predictions between the base (i.e., training) distribution and the previously unseen target distribution.
In Figure \ref{fig:main_figure}, we show how $DoC$ can be directly used to estimate the accuracy gap between base and target distributions.
Treating $DoC$ as a feature, we obtain regression models which substantially outperform all other methods and reduce predictive error by nearly half ($46\%$) across all challenging natural distribution shifts such as the ImageNet-VidRobust \cite{shankar2019image} and ImageNet-Rendition \cite{hendrycks2020many} datasets.
Our work demonstrates that it is possible to attain high quality estimates of a model’s accuracy across a variety of model architectures and types of distributions shifts.


\section{Related Work}

Our work touches on multiple related lines of research that have seen substantial activity over the past few years.
Hence we only summarize the most closely related work in this section.


\noindent\textbf{Domain Adaptation.} 
In the domain adaptation literature, distributional distances are frequently defined and subsequently minimized to improve model performance. 
Notable works in this vein define discriminative distances \cite{ganin2016domain, tzeng2017adversarial}, transport distances \cite{courty2017joint, damodaran2018deepjdot, lee2019sliced}, confusion distances \cite{tzeng2014deep}, and entropic minimizations \cite{wang2020fully, saito2019semi}. 
Some of these works attempt to bound the degradation of model performance due to distribution shift \cite{ben2006analysis, cortes2019adaptation}, based on an assumption of covariate shift \cite{shimodaira2000improving}. 
These approaches focus on measures of distribution difference, yet they 
seldom investigate questions of predicting accuracy or calibration. 
Through test-time adaptation, \cite{nado2020evaluating} have noted calibration improvements, but this comes at the expense of degraded model performance on natural distribution shifts.

\noindent\textbf{Calibration.}
Calibrating large DNNs is an active area of research with methods focused on post-hoc calibration \cite{guo2017calibration}, encoding uncertainty \cite{gal2016dropout,lakshminarayanan2017simple, wenzel2020hyperparameter, hoffman2013stochastic, riquelme2018deep}, and training interventions \cite{hendrycks2019augmix, hendrycks2018deep, hendrycks2020many} dominating the field.
However, a large-scale evaluation of calibration over natural distribution shifts  showed that these methods seldom exhibit desired calibration performance in the presence of distribution shifts \cite{ovadia2019can}. 
Research in calibration focuses on instance-level evaluation of models' predictive uncertainty, while our accuracy prediction task is focused on aggregate performance over an entire test set.

\noindent\textbf{Natural Distribution Shift.}
Several works in recent years have examined the problem of natural distribution shift, roughly divided into two camps. 
One line of work, Domain Generalization,  has focused on the problem of learning models that adapt well to various forms of distribution shift \cite{torralba2011unbiased, koh2020wilds, peng2017visda, gulrajani2020search, tobin2017domain}. 
These works establish datasets consisting of multiple distinct distributions where some of the distributions could be used for training a model and others for testing it.

\par
Another line of work, Robust Classification, has focused on assuming one training distribution and evaluating or improving performance over a set of specific types of distribution shifts \cite{shankar2019image, recht2019imagenet, hendrycks2020many, hendrycks2019natural, taori2020measuring}. We blend both lines of work, as we assume training happens with only one training distribution, yet we allow for multiple subsets of the distributions to be used for calibrating the model and evaluating performance.
Several recent works have explored the question of model evaluation with  \cite{zilly2019frechet, deng2021labels} using Frechet Distance and \cite{deng2021does} using rotation prediction to encode shift, however these approaches struggle to predict generalization gaps on natural distribution shifts. 
\cite{chen2021mandoline} explores the problem from an importance weighting and density estimation perspective, in contrast to the distribution distance and calibration based investigation done in this work.
We plan to incorporate experiments based in this work into an updated version of this paper and evaluate over the variety of shifts explored in this paper.

\noindent\textbf{Predicting Generalization Gaps in Deep Learning.}
Several recent works \cite{jiang2018predicting, jiang2019fantastic} have explored empirical and theoretical methods to predict the generalization gap of deep learning models, i.e., the gap between training and test accuracy.
These methods focus on identifying characteristics of the model and relationship between the model and training distribution, which can be used to predict the generalization gap on unseen examples in the test distribution.
Since these methods assume that the training and test distributions are the same, they do not incorporate characteristics of the test distribution and do not have a mechanism for adjusting their predicted generalization gaps under distribution shifts.
So when used in our setting with distribution shift, these methods would only predict a single accuracy regardless of the specific test set and hence underperform on the task of predicting accuracy under distribution shift.

\section{Baseline Distance Measures}
\label{sec:elsewhere}

We let $F$ represent a model and $F(x)$ represent the output probabilities of $F$ over instance x.
We are interested in estimating the accuracy of $F$ over an arbitrary target distribution $\mathcal{T}$ as computed over test samples comprising dataset $T$.
We also assume the model $F$ is trained over distribution $\mathcal{B}$, whose accuracy can be computed over samples comprising the held-out test dataset $B$.
As some of the distribution shift settings also contain changes in the label space,
we define $K_B$ as the set of labels present in $\mathcal{B}$ and $K_{B \cap T}$ as the intersection of labels present in $\mathcal{B} \text{ and } \mathcal{T}$. 
$B_{B \cap T}$ is the set of points in $B$ whose labels are in $K_{B \cap T}$. 
We are interested in predicting either the accuracy, $A_T^B = \mathbf{Acc}(T_{T \cap B})$, or the accuracy gap, $\Delta{\mathbf{Acc}}(B, T)  = \mathbf{Acc}(B_{B \cap T}) - A_T^B$, on unseen distributions.
Given access to our models accuracy over the base distribution, $A_B^T$, predicting either $A^B_T$ or $\Delta{\mathbf{Acc}(B,T)}$ would allow us to recover the other term.
Some methods directly predict accuracy while others predict the accuracy gap; this is further specified in supplementary materials.
To enable fair comparisons, we evaluate all approaches on their ability to predict accuracy and apply the $A^T_B + \Delta{\mathbf{Acc}(B,T)}$ transformation before evaluation. 
For readablity, $B^\prime$ and $T^\prime$ will be used to represent $B_{B \cap T}$ and $T_{T \cap B}$ respectively.
\par

We focus on computing a measure of distance $S_{B^\prime,T^\prime}$ between $B^\prime$ and $T^\prime$ which would allow us to use a regression model $R$ to predict $\hat{A}_T^B = R(S_{B^\prime, T^\prime})$. 
Our measures of distance, $S$, are computed based on a featurizer, $F^\prime$, associated with the classification model $F$. 
$F^\prime$ may represent any features that can be extracted from the model; most commonly these are activations from the penultimate layer of the network, but alternatively they may be logits, probabilities, or the activations from convolutional layers. 
Unless explicitly stated, all $F^\prime$ in this work deal with the penultimate layers of a deep neural network (DNN). 
Given base $B$ and target $T$ datasets, we let $S_{B,T} = M(B, T, F^\prime)$ where $M$ can be any function that returns a scalar $S_{B,T}$ based on $B, T, \text{and, } F^\prime$.
\par
The measures of distance we explore in this work are commonly used in domain adaptation literature.
A large number of works in domain adaption look into the theoretical and practical benefits of minimizing discriminative distances on new domains \cite{ganin2016domain,ben2006analysis, cortes2019adaptation, tzeng2017adversarial}.
We train domain discriminators and look at both their final AUC, and A-proxy \cite{ben2006analysis} distances on held-out test samples. 
Prior works explore Frechet distance \cite{zilly2019frechet,deng2021labels} as a useful measure for predicting accuracy and as such we include it in our analysis.
We also evaluate Maximum Mean Discrepancy (MMD) \cite{borgwardt2006integrating}, which is commonly used in domain adaptation methods \cite{tzeng2014deep} and has even been shown to be correlated with target domain accuracy \cite{tzeng2014deep, long2017deep}.
The pretext task of predicting which of 4 $(0^{\circ}, 90^{\circ}, 180^{\circ}, 270^{\circ})$ rotations has been applied to an image has been useful in self-supervised learning, adaptation, generalization, and even predicting generalization on unseen shifts. 
Our Rotation prediction approach mirrors that of concurrent work \cite{deng2021does}, however we treat the model representations as fixed and do not update weights to help solve the rotation task.
More details behind each of these approaches are provided in the supplementary material.

\noindent\textbf{Predicting Accuracy.}
In order to learn our regression model $R$ and evaluate its performance over novel shifts, we establish data groupings $C$ and $V$ for calibration and validation, respectively. 
Our regression model, $R$, is trained over the accuracy gaps $\Delta{\mathbf{Acc}}(B, T)$ and measures of difference $S$  present in the calibration grouping $C$. We train $R$ to minimize Mean Squared Error (MSE) with the accuracy gap as the target, as specified below:

\begin{equation}
  \sum_{T \in C}{\|R(S_{B,T}) - \Delta{\mathbf{Acc}}(B,T) \|^2_2} \; .
\end{equation}

We assess the quality of our regression model over the validation data grouping $V$ to understand how it generalizes to unseen shifts. 
Our primary experiments are done using linear regression to learn $R$, and the experiments with non-linear regression models in the supplementary materials follow the same trends as the linear models.
The entire prediction pipeline is illustrated in Figure \ref{fig:sec_figure}.



\section{Predicting Performance with Difference of Confidences}

A perfectly calibrated model is defined as follows with $\hat{P} = \max(F(x))$ corresponding to the predicted confidence and $\hat{Y} = \argmax(F(x))$ corresponding to predicted label:

\begin{equation}
P(\hat{Y} = Y | \hat{P} = p) = p, \; \forall \, p \in [0,1] \; .
\end{equation}

Thus for a  perfectly-calibrated model, the expected value of the model's confidence over a distribution should correspond to the model's accuracy over this distribution:
\begin{equation}
\mathbb{E}[\hat{P}] = \mathbb{E}[\hat{Y}=Y] = \mathbf{Acc}
\end{equation}

However, prior work  establishes that modern neural networks are often miscalibrated \cite{guo2017calibration,ovadia2019can}; they also show that the average confidences of these models differ from the model's accuracy over the respective distributions.
The average confidence ($AC$) for $F$ on $B$ is defined as follows: 
\begin{equation}
    AC_B = \frac{1}{|B|} \sum_{x \in B}{\max_{\{K_B\}}(F(x))} \; .
\end{equation}

Average confidence is used as a baseline to directly estimate accuracy on an unseen distribution. 
While average confidence alone proves an unreliable estimate of accuracy, we show here that the difference of confidences ($DoC$) is informative over various forms of distribution shift.  
This allows for accurate predictions of model performance on challenging domains.
As the average confidence is directly related to calibration, we also report results average confidence after performing temperature scaling on the ImageNet-Validation dataset as described in \cite{guo2017calibration} as \textit{AC TempScaling}.

\par

\label{sec:difference}

Given the definition of $M$ above, difference of confidences $DoC$ and average confidences $AC$ can be computed based on a featurization $F^\prime$ of the probabilities of the model. 
As some of the distribution shifts we explore also change the label space, we need to consider the average confidence with respect to classes present in both $T$ and $B$, as $K_{B \cap T}$.
We calculate average confidence over instances of $B^\prime$ which contains classes present in the $K_{B \cap T}$: 
\begin{equation}
    AC^T_{B} = \frac{1}{|B^{\prime}|} \sum_{x \in B^{\prime}}{\max_{\{K_{B \cap T}\}}(F(x))} \; .
\end{equation}

We propose difference of confidences $DoC$ as a way to quantify distribution shifts:
\begin{equation}
    DoC_{B,T} = AC^T_B - AC^B_T \; . 
\end{equation}
We show that this simple strategy encodes useful information about  distribution shift that can lead to successful predictions of accuracy changes. 

\par
As our proposed measure, $DoC$, is a summarizing function of the probabilities of our model $F$, we also report a variant of our method with difference of average entropy of a model's output probabilities.
Entropy may be used as a measure of uncertainty in its own right and \cite{wang2020fully} show a positive correlation between entropy and accuracy of batches of data. We define average entropy as: 
\begin{align}
\text{Ent}^T_B &= -\frac{1}{|B^{\prime}|} \sum_{x \in B^{\prime}}\big{(}{\sum_{\{K_{B \cap T} \}}F(x)log(F(x))\big{)}}
\end{align}
and difference of average entropy ($DoE$) as:  
\begin{align}
\text{DoE}_{B,T} &= \text{Ent}^T_B - \text{Ent}^B_T \; .
\end{align}

As seen below, our empirical results suggest that $DoE$ shares characteristics with $DoC$ which allows it outperform other baselines on predicting natural shifts, yet it does not perform as well as $DoC$ reducing prediction error.
We evaluate both $DoE$ and $DoC$ as features encoding the magnitude of distribution shift and show they both exceed performance of other well-known approaches described in Section \ref{sec:elsewhere}.

\section{Experiments}

\begin{table}[ht]
\centering

\rowcolors{2}{white}{gray!25}
\caption{Mean Accuracy Gaps of Distribution Groupings}
\label{tab:accs}
\begin{tabular}{lcccc}
\hline
Model & Natural & Synthetic & ImageNet-A  \\ \hline
AlexNet & 0.25 & 0.36 & 0.74  \\
VGG-19 & 0.26 & 0.42 & 0.87 \\
RN-18 & 0.24 & 0.37 & 0.86   \\
RN-34 &  0.24 & 0.35 & 0.88 \\
RN-50 & 0.25 & 0.37 & 0.918   \\ 
RN-101 & 0.23 & 0.33 & 0.88  \\ 
RN-152 & 0.22 & 0.32 & 0.87\\
RNxt-101 &  0.22 & 0.32 & 0.84   \\ 
WRN-101 & 0.22 & 0.33 & 0.86   \\ 
Deepaug & 0.217 & 0.26 & 0.89  \\ 
AM & 0.23 & 0.29 & 0.89  \\ 
AM-Deepaug & 0.20 & 0.20 & 0.88  \\ 
All Models & 0.23 & 0.33 & 0.87  \\\hline
\end{tabular}
\vspace{0.1cm}
\caption*{Shows the predicted accuracy error of the models if we rely on accuracy on the base distribution (Base Acc) to evaluate model performance.
This optimistic form of model evaluation can be referenced as a naive baseline of the error we seek to reduce throughout this work.}
\end{table}

Using DNNs trained on ImageNet ~\cite{russakovsky2015imagenet}, we evaluate our ability to predict performance on unseen distributions.
This popular large-scale dataset is frequently used as a source for models to be used on different distributions \cite{kornblith2019better, huh2016makes}, as such it is important to better understand how these models perform in the presence of distribution shift.
Using ImageNet as the base dataset, $B$, allows us to easily access several pre-trained models with various architectures, training curriculum, data augmentations, and calibrations.
Additionally, there has been significant recent work \cite{recht2019imagenet,hendrycks2019benchmarking,taori2020measuring} on assessing model performance on related datasets which introduce distribution shifts and share class labels with ImageNet.

\noindent\textbf{Natural Shifts.}
Through our empirical evaluation, we examine the impact of both natural and synthetic distribution shifts. 
For the purposes of this study, we define natural shifts to be shifts caused by how the data was collected and synthetic shifts as those that can be induced by programmatic alterations applied to the input images.
Specifically, we look at ImageNet-V2 \cite{recht2019imagenet}, ImageNet-VidRobust \cite{shankar2019image}, ImageNet-Rendition \cite{hendrycks2020many}, ImageNet-Sketch \cite{wang2019learning}.
These settings introduce various types and intensities of distribution shifts, over which it is desirable for our models to remain well-behaved. 
In addition to the above stated natural distribution shifts, we also examine performance over ImageNet-Adversarial \cite{hendrycks2019natural}, a distribution that contains natural images, yet they were collected in manner that is explicitly designed to hurt performance on ImageNet classification models. 
This dataset blends adversarial robustness and natural distribution shift, as such it is examined separately from the other natural datasets which are explained in more detail in the supplementary material. 

\noindent\textbf{Synthetic Shifts.}
 Synthetic distribution shifts provide a compelling way for us to investigate model performance under distribution shift as they are controllable in both style of perturbation and intensity of corruption. 
 Yet, since the cause of the shift in these distributions does not naturally arise from our data collection processes, there is a risk that methods which successfully tackle synthetic distribution shift may not prove useful in the more challenging and realistic natural distribution shift setting.
 We examine the common synthetic corruptions and perturbations present in ImageNet-C \cite{hendrycks2019benchmarking}.

\noindent\textbf{Approaches.}
Throughout this work we title our approaches based upon the function they use to compute distribution similarity. Frechet, Disc. AUC, Disc A-proxy, Rotation, MMD, $DoE$, and $DoE$ are the titles for approaches which predict accuracy gap, $\Delta(\mathbf{Acc})$, by training a linear regression model $R$ on top of their difference quantification $S$. The approaches $AC$, $DoC$-Feat, and $AC$ TempScaling directly estimate accuracy themselves and do not use a regression model. 

\noindent\textbf{Models.}
\label{par:models}
In order to ensure that our approach is not sensitive to certain model architectures, data augmentations, or training schemes we evaluate over a range of models with distinct training pipelines and various accuracy and robustness characteristics. Models explored include AlexNet \cite{krizhevsky2012imagenet}, VGG \cite{simonyan2014very}, ResNet \cite{he2016deep}, ResNext \cite{xie2017aggregated}, WideResNet \cite{zagoruyko2016wide}, DenseNet \cite{huang2017densely}, and Deep Ensembles \cite{lakshminarayanan2017simple}, AugMix \cite{hendrycks2019augmix}, DeepAugment \cite{hendrycks2020many}. In Table ~\ref{tab:accs} we present the accuracies of these models on the natural distribution shifts described above. 
The results we present when comparing approaches are average results across all of the aforementioned model architectures, unless specified. Details on the exact experimental setup can be found in supplementary material; including architectural variants and training schemes for models and calibration. 

\noindent\textbf{Data Groupings.}
We take special care to ensure that our ability to predict performance on unseen distributions is not corrupted by any form of information leakage \cite{kaufman2012leakage}.
Ordering our distribution shifts into three distinct groups, two synthetic and one natural, allows us to minimize shared information between data groups and ensure that the unseen shifts in our validation group are unseen in both style and magnitude of shift. 
Some of our target distribution shifts share similarities with one another which could advantage an accuracy predictor which was exposed to similar distribution shift at calibration time.
To mitigate this we place all shifts with known similarities into the same data group, so that our validation shifts share little in common with our calibration shifts and gives a more faithful measure of an approach's ability to generalize.
\par
Our work's chief concern is the ability to predict natural distribution shifts, and as such focuses on settings with synthetic calibration and natural validation datasets.
We detail the types of shifts and corruptions contained in each grouping in the supplementary material and present ablation studies on the impact of different calibration groupings there as well.

\subsection{Results}

Learning models that transfer well from synthetic to natural environments (Syn2Real) \cite{richter2016playing,peng2018syn2real, dosovitskiy2017carla} is appealing due to the relative ease of creating more synthetic data.
As such we conduct the majority of our experiments in challenging setting of calibrating over synthetic shifts and predicting performance on natural distribution shifts.


\begin{figure}[ht]
\centering\includegraphics[width=0.5\textwidth]{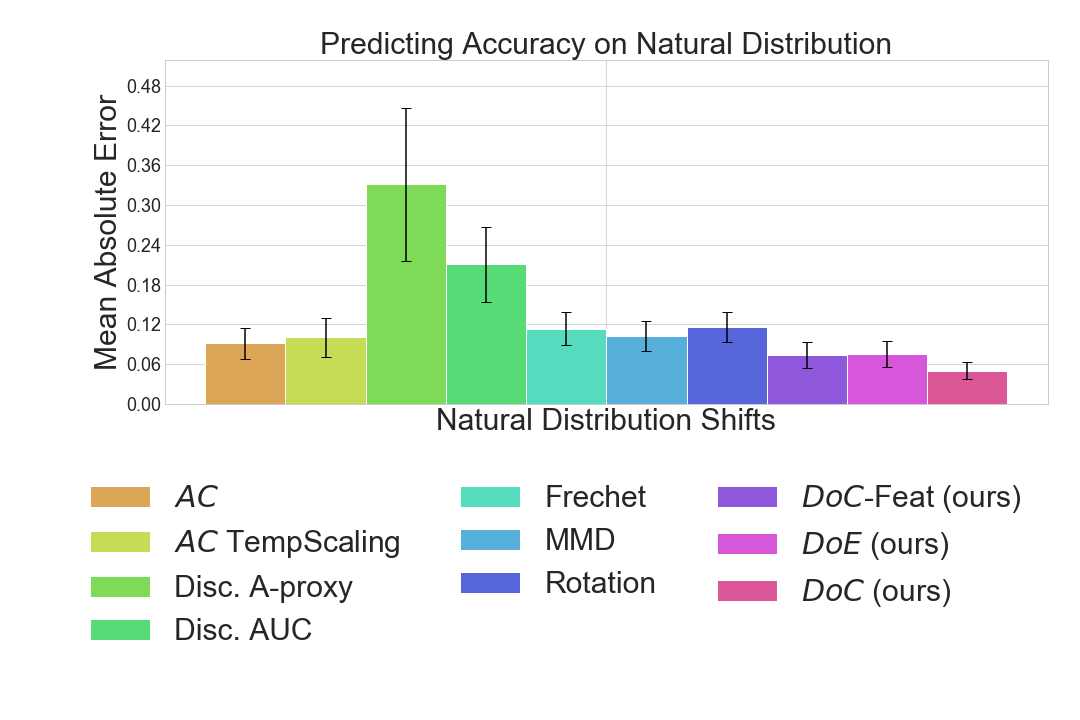}
\caption{
On the challenging and important task of predicting performance over natural distribution shifts with exposure only to synthetic shifts, our approach $DoC$ outperforms all alternatives on predicting accuracy. 
Traditional distance metrics, such as Frechet distance and A-Proxy distance perform worse than the $AC$ baseline when calibrated over synthetic shifts.}
\label{fig:arch_natural}
\end{figure}

\noindent\textbf{Synthetic to Natural.}
In Figure \ref{fig:arch_natural}, we present the results of 10 different approaches for predicting model accuracy over 6 different natural distribution shifts with 12 distinct neural network architectures.
We visualize the results with mean absolute error (MAE) and confidence intervals in the main paper, and present scatter plots of the data points in the supplementary materials.
Across these various distribution shifts and ImageNet models, we find that the three approaches proposed in our work, ($DoC$, $DoC$-Feat, and $DoE$), are not only the best performing approaches but are also the only approaches to outperform the $AC$ baseline.
With a MAE of $0.092 \pm 0.019$, $AC$ outperforms most prior measures of difference explored in this work with only MMD ($0.10 \pm 0.018$) having substantial overlap.  
Our best performing approach, $DoC$, reduces MAE by $>45\%$, with an average error in accuracy prediction of only $5.0 \pm 0.010$.
$DoE$ and $DoC$-Feat perform comparably to one another, $0.075 \pm 0.015$ and $0.072 \pm 0.016$ respectively, despite $DoE$ leveraging the additional information from the calibration datasets and $DoC$-Feat not using any regression model.
Interestingly, though $DoE$ and $DoC$ operate over the same featurization $F^\prime$, output probabilities, $DoC$ significantly outperforms its entropy-based counterpart through discarding all non-maximum probabilities.
Table \ref{tab:accs} shows a MAE of $0.23$ over the natural data grouping if one assumes there was no accuracy gap or equivalently estimates target accuracy from Base Acc.
When calibrated over synthetic distribution shifts, A-proxy \cite{ganin2016domain} and Discriminative AUC do not significantly improve on this naive baseline.

\par
In addition to understanding which approaches work best overall, we are also interested in understanding which, if any, situations our approach fails to improve performance. 

\begin{figure}[ht]
\centering\includegraphics[width=0.5\textwidth]{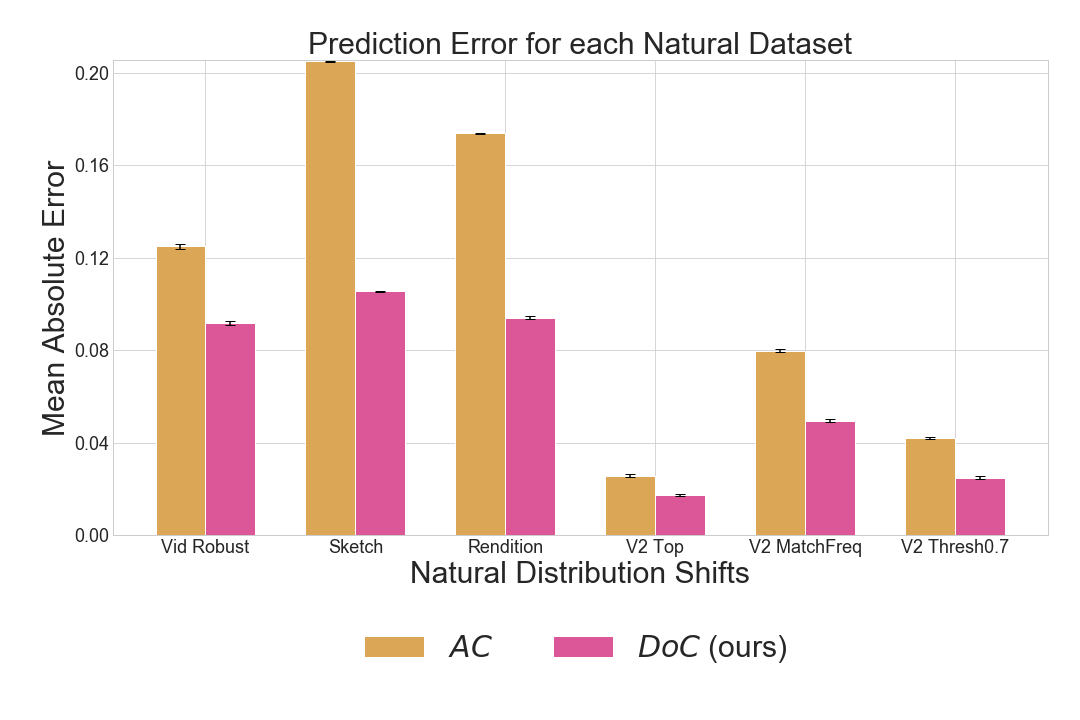}
\caption{Expanding the results of figure \ref{fig:arch_natural} we observe that baseline accuracy $AC$ varies significantly over different natural distribution shifts, with datasets not exclusively composed of natural photography images producing the highest error. However, $DoC$ significantly reduces error across each shift by nearly $50\%$.
}
\label{fig:easy-nat}
\end{figure}

In Figure \ref{fig:easy-nat}, we examine average error over all models for the baseline $AC$ and our best performing approach $DoC$ over each dataset of natural distribution shift.
We see that for some distributions, \textit{Vid Robust} and \textit{V2 Top}, $AC$ does a good job at predicting model accuracy with $<5\%$ error. 
Other distribution shifts, ImageNet-Rendition and ImageNet-Sketch, the baseline has a much higher magnitude of error.
Encouragingly our approach $DoC$ reduces error by close to $50\%$ across each form of natural distribution shift. 
The results from Figure \ref{fig:easy-nat} indicate that it is harder to predict accuracy in the face of certain forms of distribution shift; even after calibration the best average error for ImageNet-Sketch, and ImageNet-Rendition datasets is still higher than the baseline $AC$ error on the remaining distributions. 
It is worth noting that not all images in ImageNet-Rendition and Sketch are natural photos which may present a more challenging form of shift than the other datasets which are comprised exclusively of natural photography.
\par

\begin{figure}[ht]
\centering\includegraphics[width=0.5\textwidth]{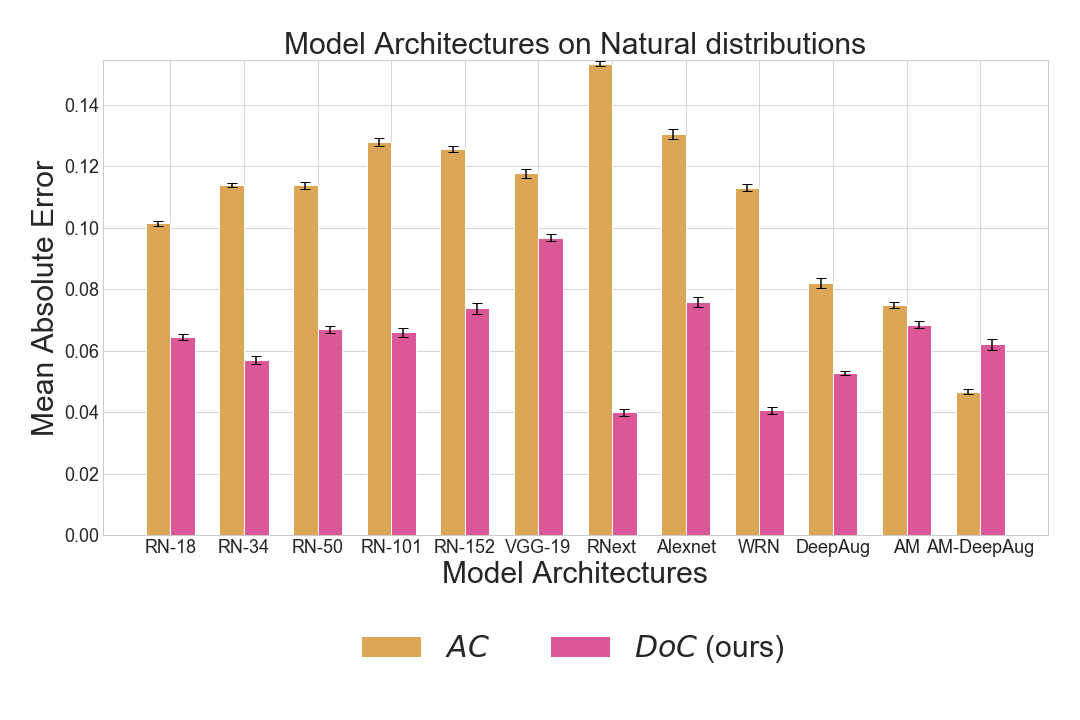}
\caption{When evaluated over natural distribution shifts $DoC$ consistently improve performance over each specific model type, with the exception of those trained using Augmix. Models trained with Augmix are designed to combat synthetic corruptions thus calibrating on synthetic corruptions may harm results.  
}
\label{fig:imagenet-models-perf}
\end{figure}

\noindent\textbf{Model Specific Performance.}
In Figure \ref{fig:imagenet-models-perf}, we examine how our approach performs on natural distributions for each model architecture and see that $DoC$ improves performance for every model variant except Augmix-DeepAugment (AM-DeepAug).
Furthermore, we observe a steep decline in the magnitude of improvement offered from our $DoC$ approach as it is evaluated over models which incorporate synthetic corruptions into the training process (Augmix, AM-DeepAugment). 
As we do not replicated the corruptions present in the Augmix training process when comparing between the $B$ base dataset and the calibration datasets $C$, it is likely that our measured distances with respect to these models are off as a result.
For all models not trained with these robustness interventions, our $DoC$ significantly improves the ability to predict accuracy over unseen natural distribution shifts.

\begin{figure}[ht]
\centering\includegraphics[width=0.5\textwidth]{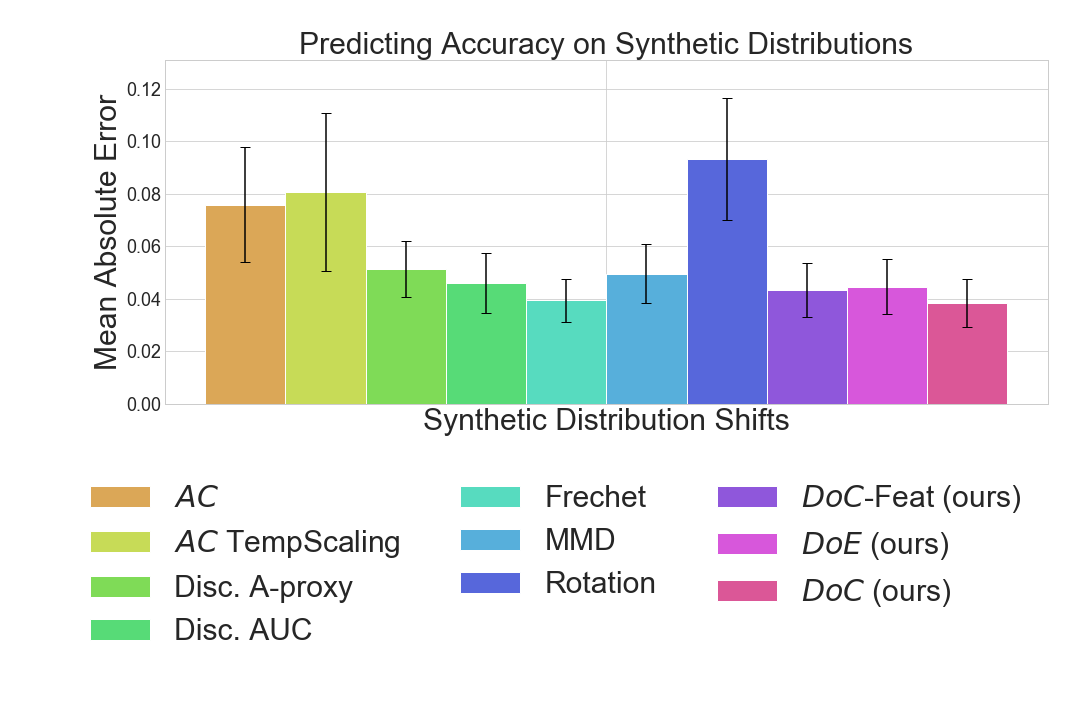}
\caption{
When calibrated with exposure to one set of synthetic shifts, all of the approaches substantially reduce the base accuracy MAE of 0.33 shown in Table \ref{tab:accs}, when predicting on a held out set of synthetic shifts.
Furthermore, with the exception of rotation prediction, all approaches outperform the $AC$ baseline for predicting performance under synthetic distribution shifts. 
While, $DoC$ and Frechet distance are the best performing approaches in this setting, most approaches are similarly able to predict accuracy over these synthetic shifts.
}
\label{fig:syn2-perf}
\end{figure}

\noindent\textbf{Synthetic to Synthetic.}
Using the same setup as in the natural to synthetic analysis, in Figure \ref{fig:syn2-perf} we evaluate our approaches over novel synthetic shifts. 
Unlike the synthetic to natural setting all approaches studied substantially improve on the $0.33$ MAE of the base accuracy method shown in Table \ref{tab:accs}.
Furthermore, most approaches outperform the baseline of $AC$, MAE ($0.076 \pm 0.017$), with the sole exception being Rotation Prediction, MAE ($0.093 \pm 0.019$).
Frechet Distance and $DoC$ are the best performing approaches with near identical MAE of $0.039 \pm 0.007$. 
These best performing approaches yield a $49\%$ reduction in relative error from the $AC$ baseline and an $88\%$ reduction in error from the base accuracy method.

As the majority of the distance metrics explored in this work overlap significantly on the synthetic to synthetic prediction task and noticeably reduce errors in predicted accuracy, studying this form of shift alone may not lead to the insight that $DoC$ methods generalize much better than their competitors when the style of distribution shift changes radically.
The observation that metrics  which reduce error for synthetic shifts, may have an inverse relationship with natural distribution shifts is worth highlighting and better understanding, as it could either indicate that these methods primarily encode information useful for synthetic shifts, or that the calibrating on synthetic data does not allow them to learn useful mappings over natural distribution settings. 


\par
While $DoC$ improves upon the $AC$ baseline for each form of natural distribution shift, we note that in 2 of the 8 synthetic shifts explored (Defocus Blur and Gaussian Blur) $DoC$ actually decreases the accuracy of the predictions. These results are presented in the supplementary material and merit further investigation. 

\begin{figure}[ht]
\centering\includegraphics[width=0.5\textwidth]{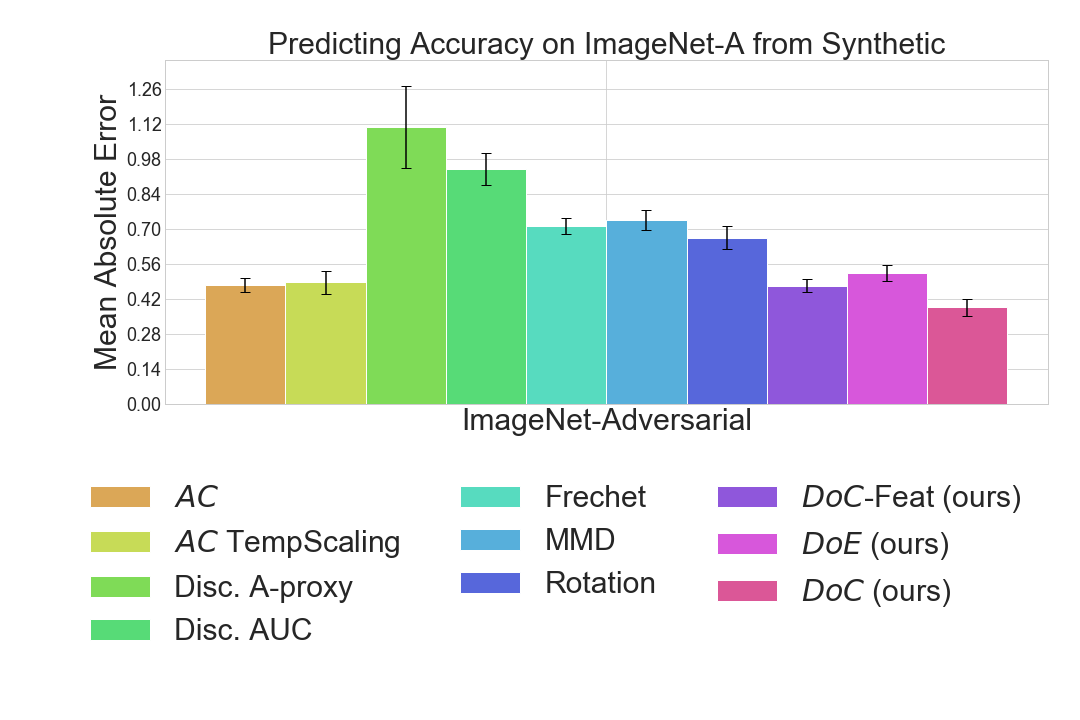}
\caption{
As ImageNet-A was adversarially designed, it presents a distinctly challenging scenario for predicting model accuracy.
$DoC$ is the only approach to significantly improve upon the $AC$ baseline when calibrated over synthetic shifts, yet it still much room to improve with a high MAE of $0.389 \pm 0.027$.
}
\label{fig:imagenet-A}
\end{figure}

\noindent\textbf{Adversarial Distribution Shift.}
ImageNet-A \cite{hendrycks2019natural} produces a uniquely challenging scenario for our approaches. 
The dataset was designed in an adversarial manner with knowledge of the predictions of an ImageNet classifier, however, unlike many other adversarial tasks, it is composed of non-perturbed images. 
In Figure \ref{fig:imagenet-A}, we observe that this distribution shift produces the highest predicted accuracy error of shifts we have studied across all approaches. 
In this setting, our approach $DoC$, MAE $0.389 \pm 0.027$, is the only approach able to noticeably reduce error from the $AC$ baseline, MAE $0.476 \pm 0.024$. 
Whereas $DoE$ and $DoC$-Feat perform comparably with the $AC$ baseline, all of the other approaches significantly increase predicted error on this task. 
\par
\begin{figure}[ht]
\centering\includegraphics[width=0.5\textwidth]{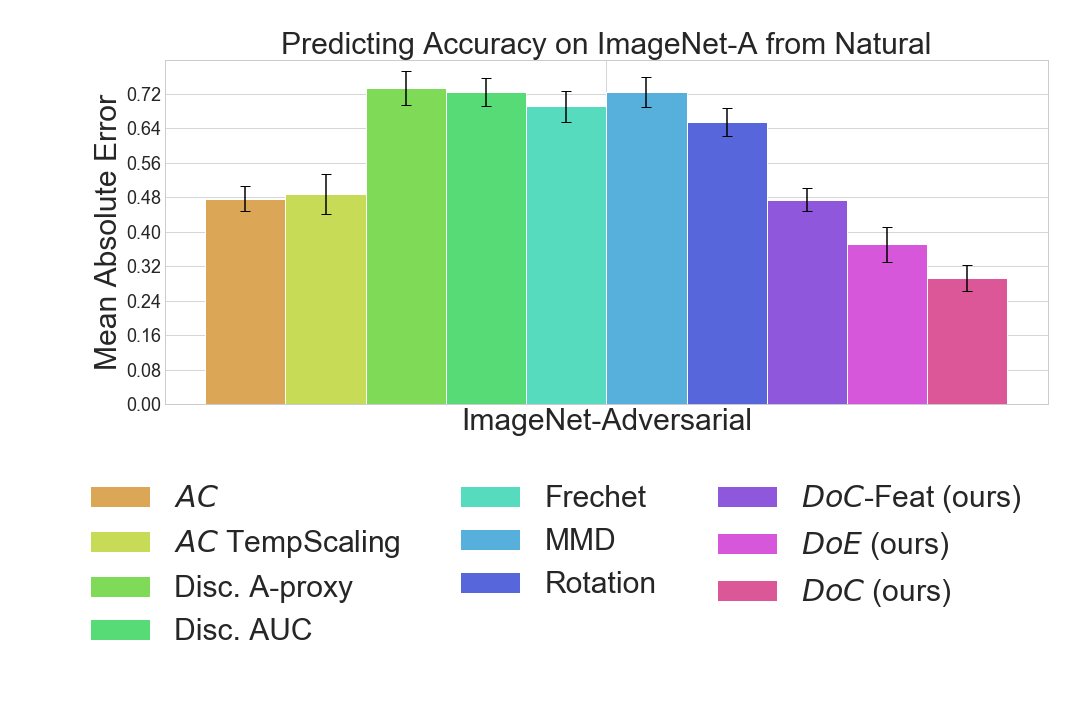}
\caption{
Calibrating over natural distribution shifts improves the performance of all approaches of their ability to predict ImageNet-A accuracy compared to Figure \ref{fig:imagenet-A}. $DoE$ and $DoC$ now improve upon baseline $AC$ performance and improve on their prior performance by $29\%$ and $25\%$ respectively, highlighting the impact of calibration groupings on predicting accuracy.
}
\vspace{-0.5cm}
\label{fig:imagenet-A-Nat}
\end{figure}

In Figure \ref{fig:imagenet-A-Nat}, we look at the results of the various approaches when calibrated over the set of natural distribution shifts.
In this setting, our results show $DoE\  \text{and}\  DoC$ are the only encodings which outperform the $AC$ baseline, with errors of $0.371 \pm 0.033$ and $0.293 \pm 0.024$ respectively.
When comparing these results to those shown in Figure \ref{fig:imagenet-A}, we see that both $DoC$ and $DoE$ noticeably improve in their ability to predict accuracy on this distinctly challenging form of distribution shift.
While all approaches which use the natural calibration make gains in their prediction accuracy, they still are significantly worse than the $AC$ baseline.
This both illustrates the importance of calibrating and evaluating over natural distribution shifts and highlights some limitations of these approaches at encoding this difference. 

\begin{figure}[ht]
\centering\includegraphics[width=0.5\textwidth]{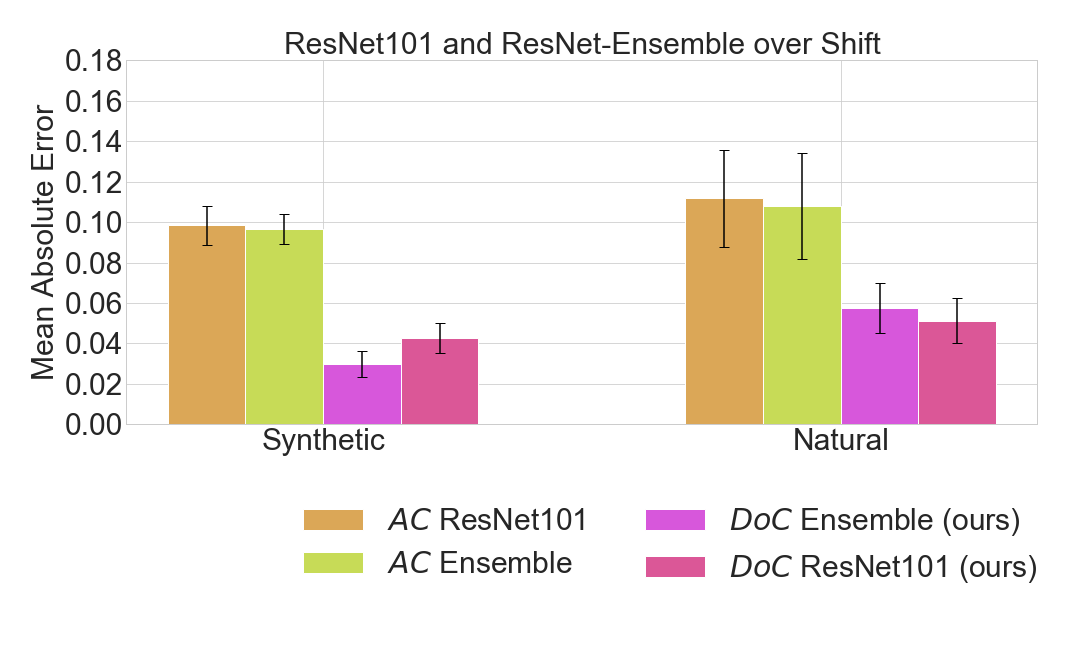}
\caption{We ompare our method $DoC$ approaches with ResNet-Ensemble, established in \cite{ovadia2019can} as best performing over distribution shifts. 
ResNet101 is compared with ResNet-Ensemble over natural and synthetic distribution shifts and we show that $DoC$ based approaches on a simple Resnet-101 outperform a deep ensemble.}
\label{fig:rn101vensemble}
\vspace{-0.4cm}
\end{figure}

\noindent\textbf{Resnets vs Deep Ensembles.}
Large-scale studies on predictive uncertainty with regards to distribution shifts \cite{ovadia2019can} have identified Deep Ensembles \cite{lakshminarayanan2017simple} as the state of the art in this space. 
In Figure \ref{fig:rn101vensemble}, we show that our post-hoc calibration approach, $DoC$, can enable a single ResNet-101 model to outperform the more costly Res-Ensemble model, defined in supplementary materials, at predicting accuracy on natural and synthetic unseen distributions. 
We also show that $DoC$ can be applied to Deep Ensembles to further improve their performance.

\section{Conclusion}
We study the problem of predicting performance change under distribution shift.
We find that a simple method, difference of classifier confidences ($DoC$), accurately predicts the performance change on a wide array of both natural and synthetic distribution shifts. The simplicity of $DoC$ and variants introduced in this paper presents a sobering view on the problem of accuracy prediction, echoing the findings of \cite{ovadia2019can}: many methods explicitly designed for uncertainty prediction seem to fall short of simpler baselines. 

While our results present a promising step forward for \textit{detecting} performance drop under distribution shift, we note that the problem is far from solved. Furthermore, the issue of \textit{reducing} the performance drop remains completely open. In this vein we highlight one avenue of future research: 

\noindent\textbf{Robust Dataset Construction.} Recent work has shown that models trained on orders of magnitude more data can make significant gains in robustness to distribution shift \cite{radford2021learning, xie2020self, taori2020measuring}.
Constructing such large datasets can be difficult and expensive, but $DoC$ could act as a potential filtering mechanism for focusing data collection on difficult sub-distributions.   

We hope that our findings can lay the groundwork for future research on both detecting and reducing performance drops induced by distribution shifts.

\noindent\textbf{Acknowledgements} This work was supported in part by DoD including DARPA’s XAI, and LwLL programs, as well as Berkeley’s BAIR industrial alliance programs.


{\small
\bibliographystyle{ieee_fullname}
\bibliography{main}
}



\include{supp}

\end{document}

%% file: supp.tex





\clearpage
\onecolumn
\section*{\centering SUPPLEMENTARY MATERIAL \hfil} 
\vskip -0.1in

\section{Non-Linear Regression}
\label{sec:nonlin}
\begin{figure}[ht]
\centering\includegraphics[width=0.6\textwidth]{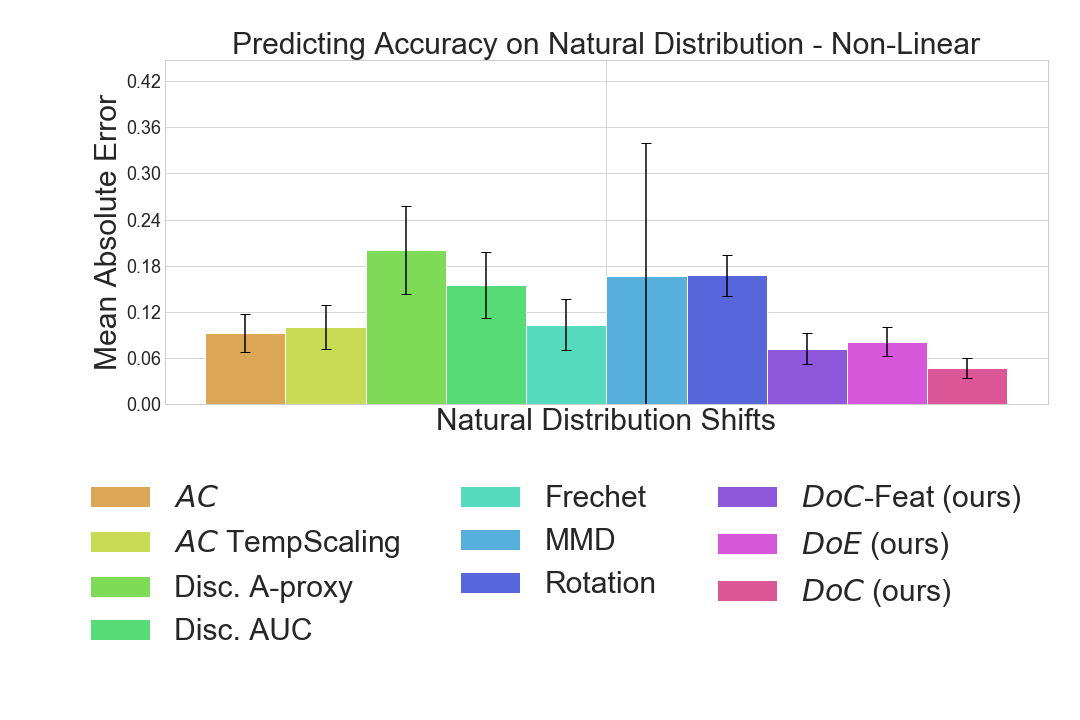}
\caption{Our approach, $DoC$, is the best performing approach when calibrating on synthetic distributions and evaluating on natural distribution shifts. 
We show that our approaches $DoC$, $DoC$-Feat, and $DoE$ are the only ones to outperform the baseline of $AC$. 
These are consistent with our findings for linear regression.}
\label{fig:nonlin-nat}
\end{figure}

\begin{figure}[ht]
\centering\includegraphics[width=0.6\textwidth]{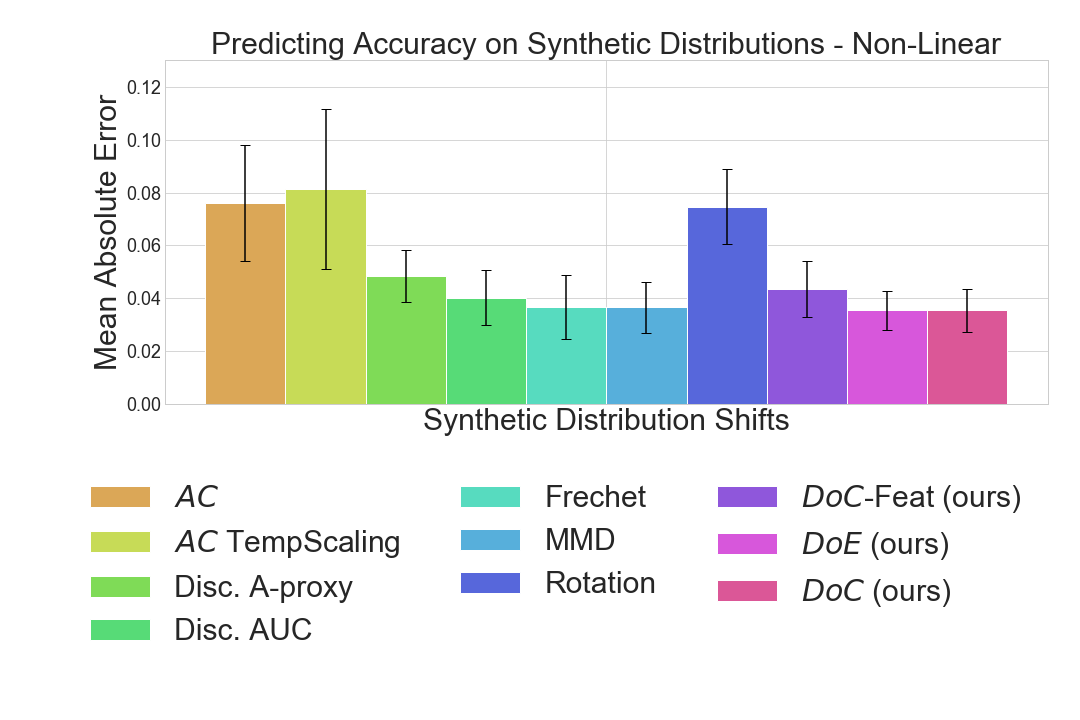}
\caption{Using the non-linear regression approach described in section \ref{sec:nonlin}, we see that all of the approaches outperform the $AC$ baseline for predicting performance under synthetic distribution shifts. 
When predicting accuracy with non-linear regression, $DoC$, $DoE$, Frechet, and MMD all perform comparably at predicting accuracy under synthetic shifts.}
\label{fig:nonlin-syn}
\end{figure}

\begin{figure}[ht]
\centering\includegraphics[width=0.6\textwidth]{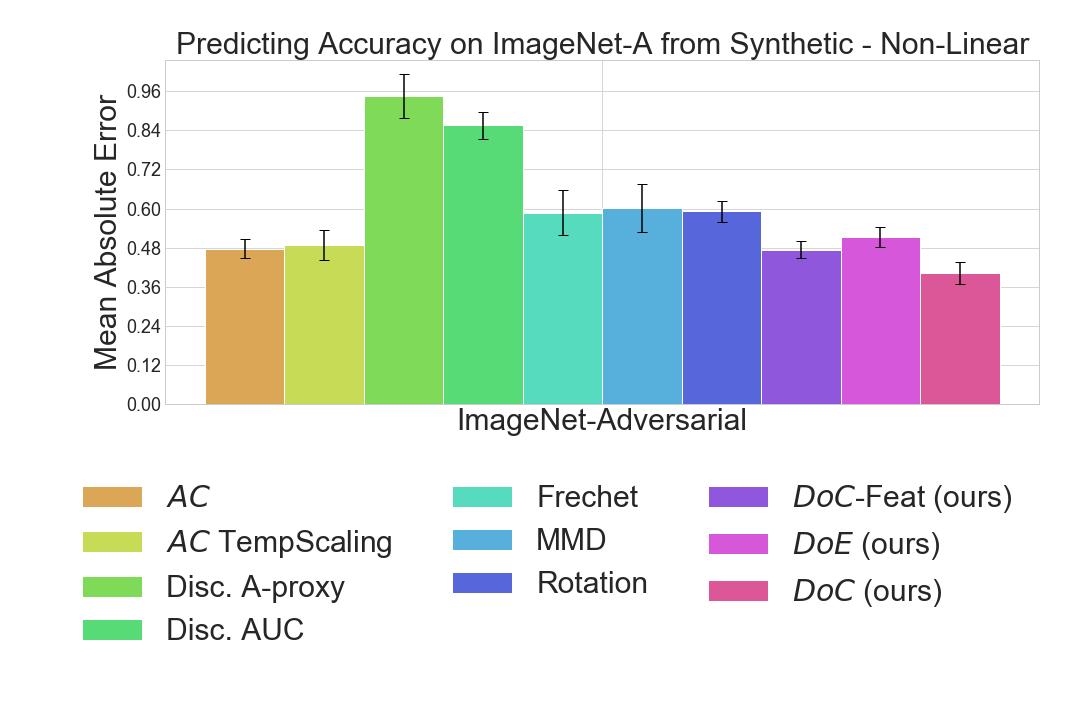}
\caption{Predicting on ImageNet-A from synthetic shifts with non-linear regression follows the same trends as with linear regression, with only $DoC$ outperforming the $AC$ baseline.}
\label{fig:nonlin-a}
\end{figure}

As referenced in section 3, we run non-linear regression experiments in which we train a Deep Neural Network to predict model accuracy given measures of distributional difference $S$. 
We use fully-connected architectures with a 3-layer fully connected architecture, with layer sizes of 512, 256, and 128 units respectively. 
The models are trained using stochastic gradient descent for 20k epochs or until convergence, with a learning rate of $1e-4$, weight decay of $1e-3$, and momentum of $0.9$. We evaluate over the same models as used in our linear analysis (Resnet-18, Resnet-34, Resnet-50, Resnet-101, Resnet-152, VGG-19, AlexNet, Resnext-101, WideResnet-101, Augmix, DeepAugment, and AM-DeepAugment). 
The results reported in Figures \ref{fig:nonlin-nat}, \ref{fig:nonlin-syn}, and \ref{fig:nonlin-a}, mirror those of the linear regression experiments. 
One noticeable difference is that the prediction error for $DoE$ on synthetic shifts decreases to equivalence with $DoC$.

\section{Distribution Shifts and Grouping}
\begin{figure}[ht]
\centering\includegraphics[width=0.6\textwidth]{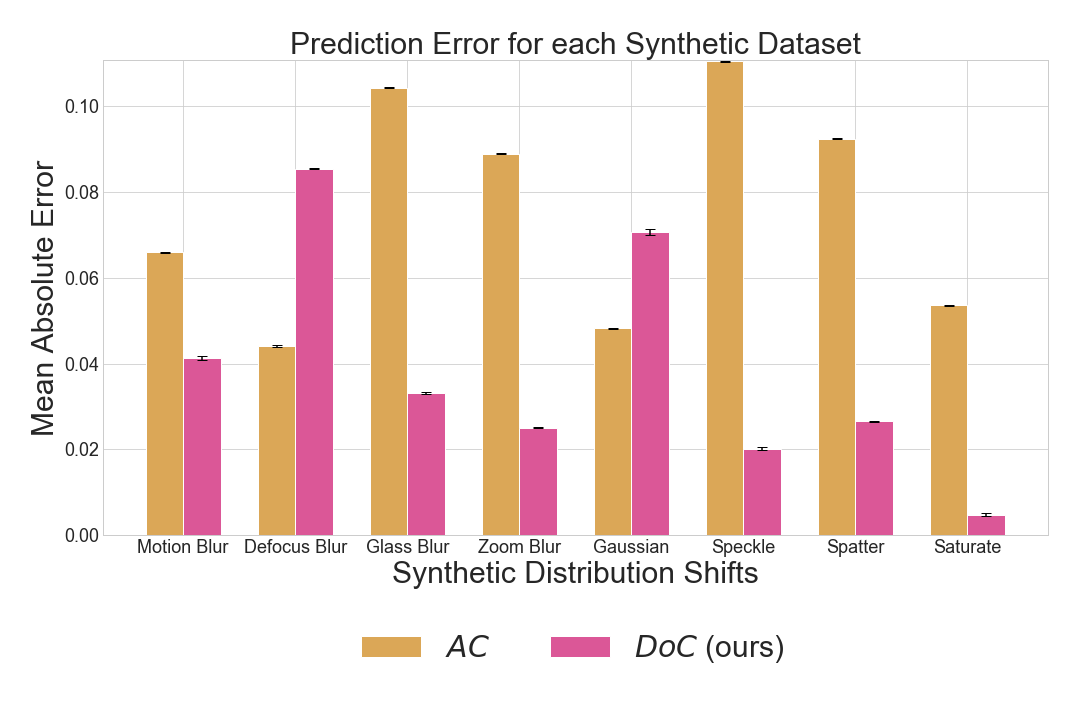}
\caption{As with  distributions, some synthetic distributions present a harder challenge than others. 
$DoC$ improves performance on 6 of the 8 corruptions but degrades performance on Defocus Blur and Gaussian Blur shifts.}
\label{fig:syn1-distributions}
\end{figure}

Among the distribution shifts, we differentiate between natural and synthetic distribution shifts. 
Synthetic distribution shifts can be generated in an automated fashion from existing data. 
We explore the synthetic shifts from the ImageNet-C dataset \cite{hendrycks2019benchmarking} and combine results from all five intensities. 
We separate the synthetic data shifts into two groupings $Syn1$ and $Syn2$. $Syn1$ is comprised of Digital shifts (Jpeg, Elastic, Contrast, and Pixelate), Noise shifts (Gaussian, Impulse, Shot), and Weather shifts (Frost, Fog, Snow, and Brightness). 
$Syn2$ is comprised of Blur shifts (Motion, Defocus, Glass, and Zoom), and Extra shifts (Gaussian Blur, Speckle, Spatter, and Saturate).
We look into how predictive performance changes for each type of synthetic shift in Figure \ref{fig:syn1-distributions}. 
While on aggregate our approach $DoC$ outperforms all other explored approaches, we see that for certain forms of synthetic shift our approach decreases the accuracy of the predictions. 
Further exploration is merited in order to best understand the cause of degradation on these specific synthetic corruptions which are both forms of image blurring (Defocus Blur, and Gaussian Blur).
\par
\begin{figure}[ht]
\centering\includegraphics[width=0.6\textwidth]{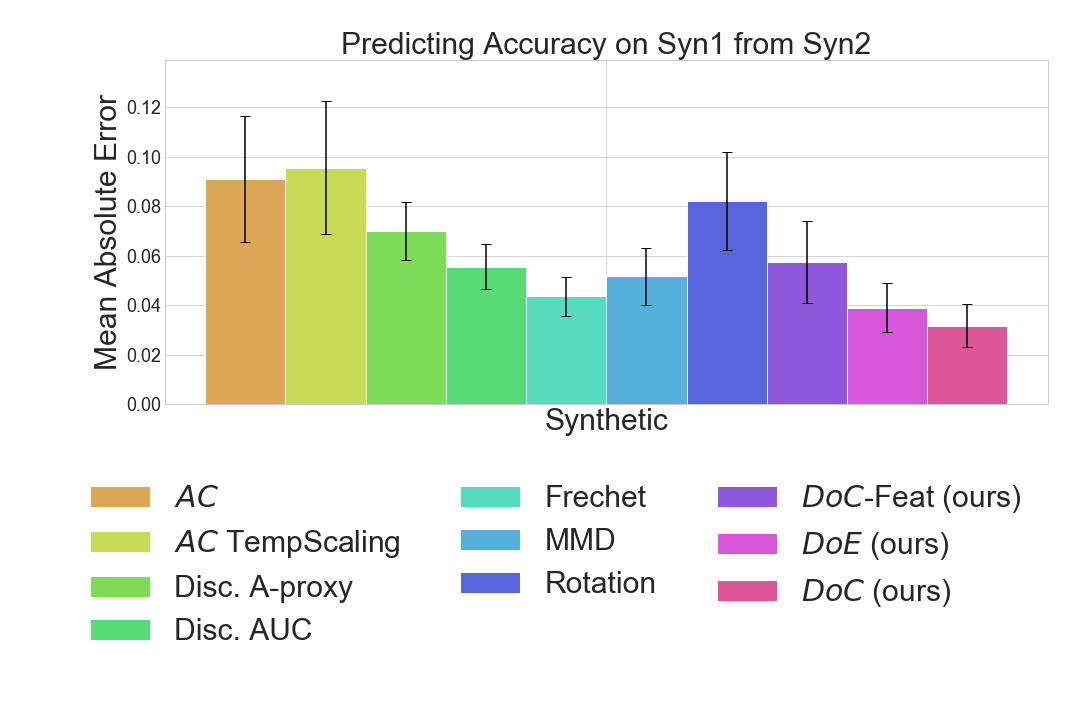}
\caption{All of the approaches outperform the $AC$ baseline for predicting performance under synthetic distribution shifts when calibrated with $Syn2$ grouping.
Our $DoC$ and $DoE$ approaches outperform the other encodings of distributional difference, with $DoC$ producing the best estimates.}
\label{fig:syn1fromsyn2}
\end{figure}
\begin{figure}[ht]
\centering\includegraphics[width=0.6\textwidth]{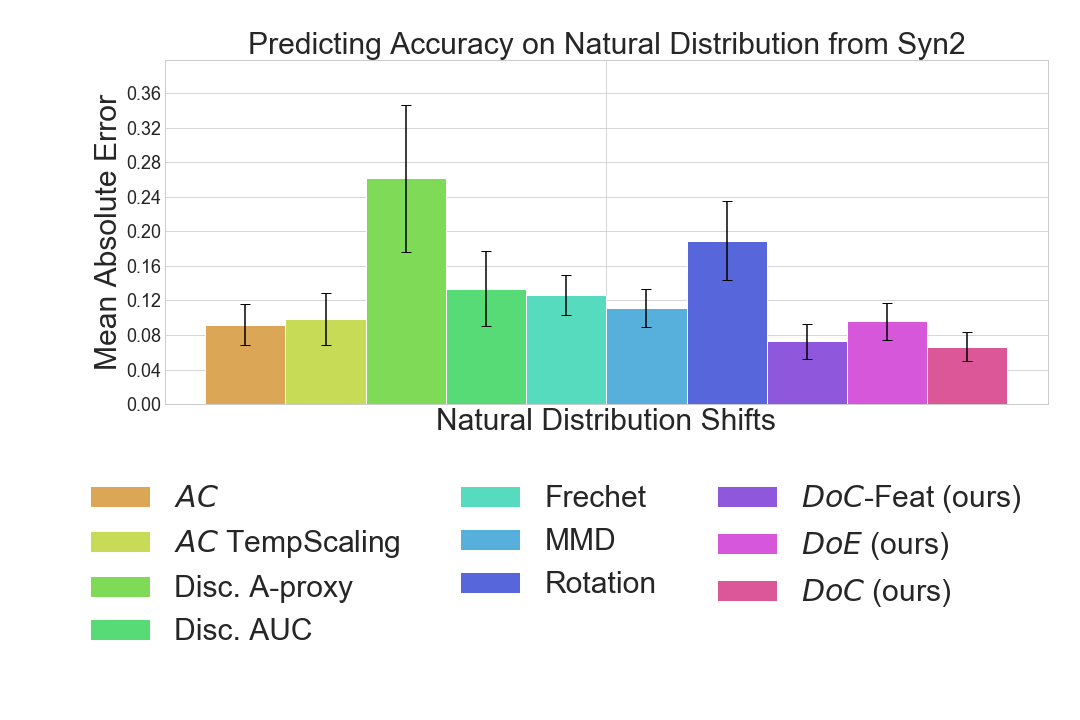}
\caption{When calibrating over $Syn2$ and evaluating on natural distributions, $DoC$ is still the best performing approach. 
$DoE$ performs near parity with our $AC$ baseline and all prior approaches performing worse than the baseline.}
\label{fig:natfromsyn2}
\end{figure}
\begin{figure}[ht]
\centering\includegraphics[width=0.6\textwidth]{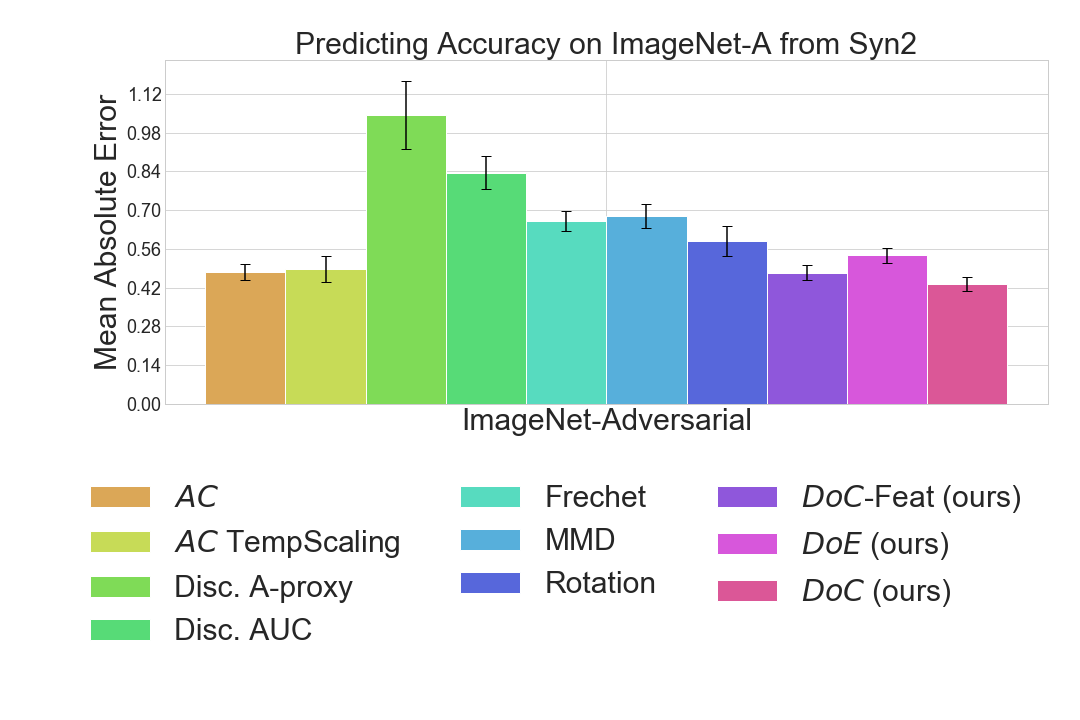}
\caption{$DoC$ is the best performing approach over ImageNet-A when calibrated with $Syn2$. 
Unlike prior synthetic calibration, $DoE$ performs worse than the baseline in this instance.}
\label{fig:imgAfromsyn2}
\end{figure}

Throughout the main text all results with synthetic calibration were calibrated over $Syn1$.  
For completeness we present the results of calibration on $Syn2$ in these Figures \ref{fig:natfromsyn2}, \ref{fig:syn1fromsyn2}, and \ref{fig:imgAfromsyn2}.
The results we observe follow the same patterns as those in the main text when calibrated over $Syn1$. 

\par
The natural distribution shifts that we explore include the three variations of ImageNet-V2 \cite{recht2019imagenet} (Top Images, Threshold-0.7, and Matched Frequency) and ImageNet-Vid-Robust \cite{shankar2019image}, both of which consist entirely of natural photography images, with ImageNet-V2 comprising the entire 1k classes of the original ImageNet and ImageNet-Vid-Robust representing a subset of ~30 classes. 
We also examine ImageNet-Sketch and ImageNet-Rendition, both of which contain object representations that are not derived from natural photography and are over a set of 200 classes contained in ImageNet. 
The aforementioned datasets represent our natural distribution data grouping, which is used for both evaluating and calibrating our accuracy predictors. 
We also evaluate over ImageNet-Adversarial ($Adv.$), but do not include this distribution in our calibrations, as it is collected in an adversarial fashion and may display different characteristics than model-agnostic shifts.
\par
As referenced in section 5, the results over all algorithmic approaches and all calibration and evaluation splits are reported in Table \ref{tab:main}. 
We observe that our approaches $(DoC, DoE)$ outperforms all other approaches in all synthetic to natural calibration settings ($Syn1 \rightarrow{Natural}, Syn1\rightarrow{Adv.}, Syn2 \rightarrow{Natural}, Syn2 \rightarrow{Adv.}$).
The results also show that $(DoC, DoE)$ outperform other approaches in natural to natural calibration settings ($Natural \rightarrow{Adv.}$).

\section{Visualizing Features and Predictions}
\begin{figure}[ht]
\centering\includegraphics[width=0.8\textwidth]{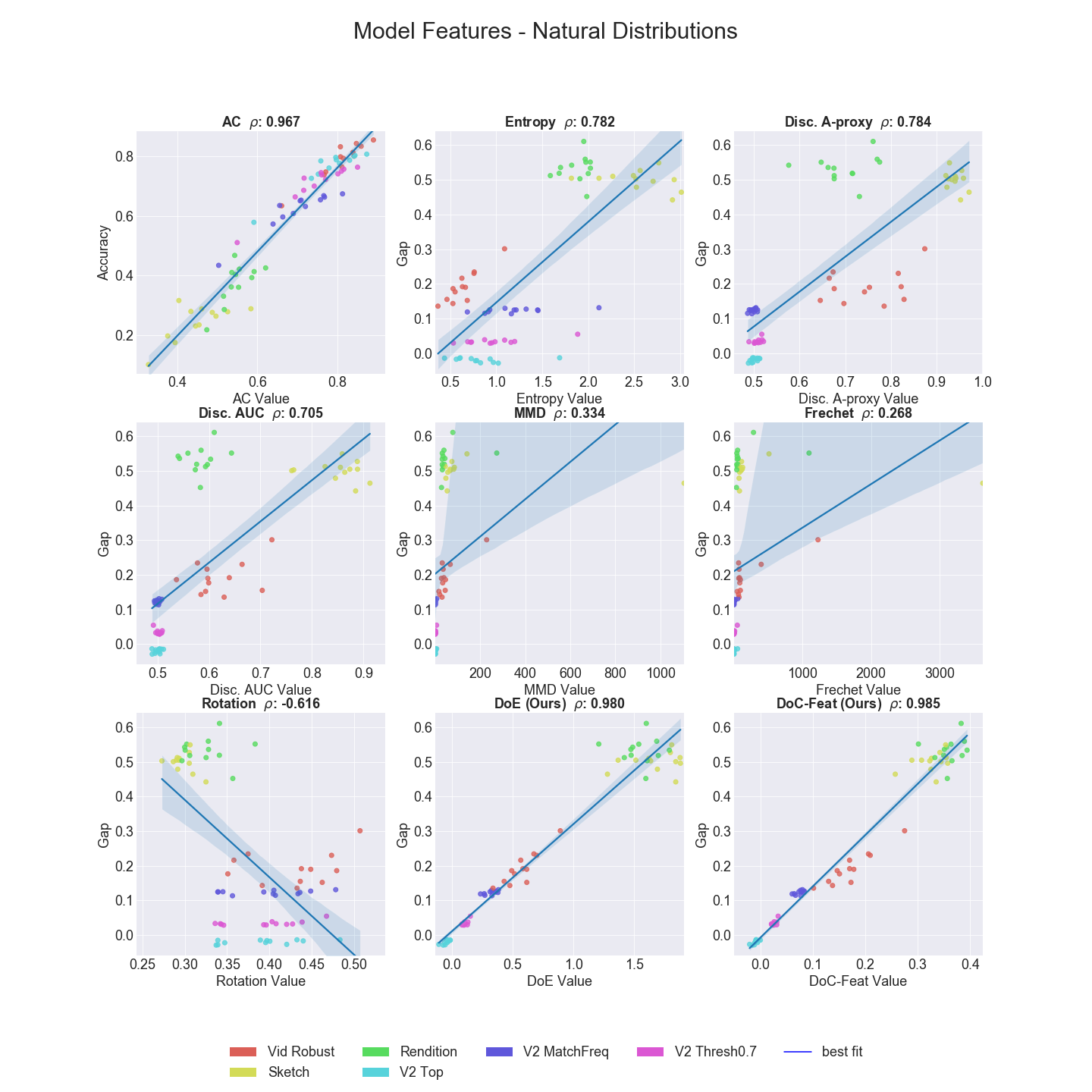}
\caption{
We plot the featurization value ($S$) and the corresponding accuracy gap for each approach explored in this work, as well as the line of best fit for all natural distribution shifts.
We also include Pearson correlation coefficient ($\rho$), for each plot to capture a measure of how informative each feature is in a global setting. 
For $AC$, we plot the accuracy instead of the accuracy gap because it is a direct estimate.
}
\label{fig:scatter-nat-raw}
\end{figure}

In figure \ref{fig:scatter-nat-raw}, we plot the values of the featurization, $S$, of each of the approaches explored in this work and the actual accuracy gap for each model and natural distribution shift.
We also show the Pearson correlation coefficient $\rho$, and the line of best for these data points to illustrate how well each of these approaches might be able to encode shifts. 
Confidence-based measures ($AC$, $DoC$-Feat, $DoE$, Entropy) display the highest levels of correlation and perhaps counter-intuitively, the discriminative discrepancy-based measures (Disc. AUC, Disc. A-proxy) show the next highest levels of correlation.
Despite, showing the second highest levels of correlation, the discriminative discrepancy-based measures were the worst performing on predicting accuracy changes over natural distribution shifts.
As reason for this that the lines of best fit and Pearson correlation coefficient are determined with knowledge of the every models' accuracy on each target distribution and best-case setting for regression model based on these features.
As our regression models are fit with exposure only to synthetic shifts and the accuracy data of one model at a time, we see that our discriminative distance predictions overfit to the synthetic shifts.
\par
\begin{figure}[ht]
\centering\includegraphics[width=0.8\textwidth]{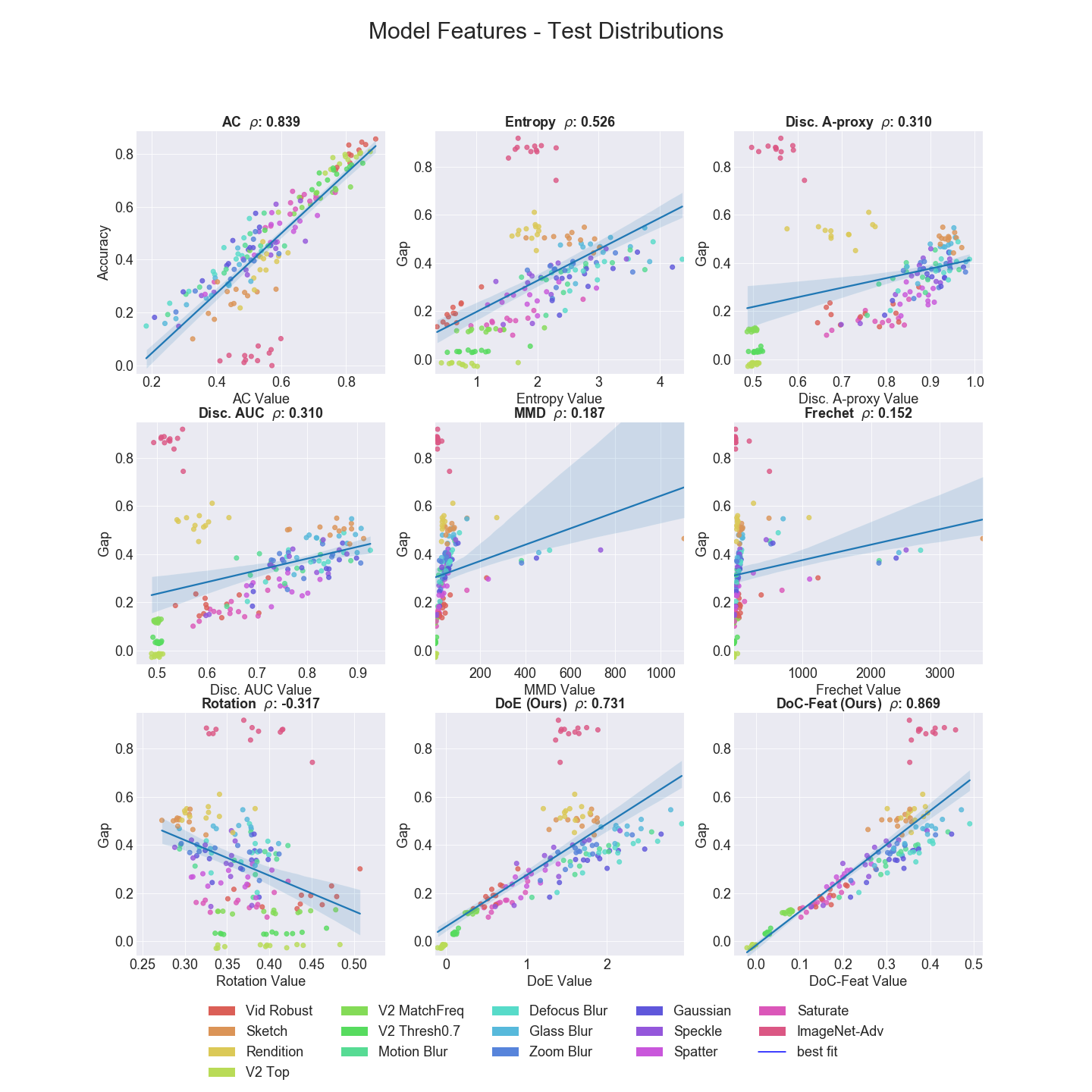}
\caption{
We plot the featurization value ($S$) and the corresponding accuracy gap for each approach explored in this work, as well as the line of best fit for all test distribution shifts ($Natural$, $Syn2$, $Adv.$).
We also include Pearson correlation coefficient ($\rho$), for each plot to capture a measure of how informative each feature is in a global setting. 
For $AC$, we plot the accuracy instead of the accuracy gap because it is a direct estimate.
}
\label{fig:scatter-all-raw}
\end{figure}
\begin{figure}[ht]
\centering\includegraphics[width=0.8\textwidth]{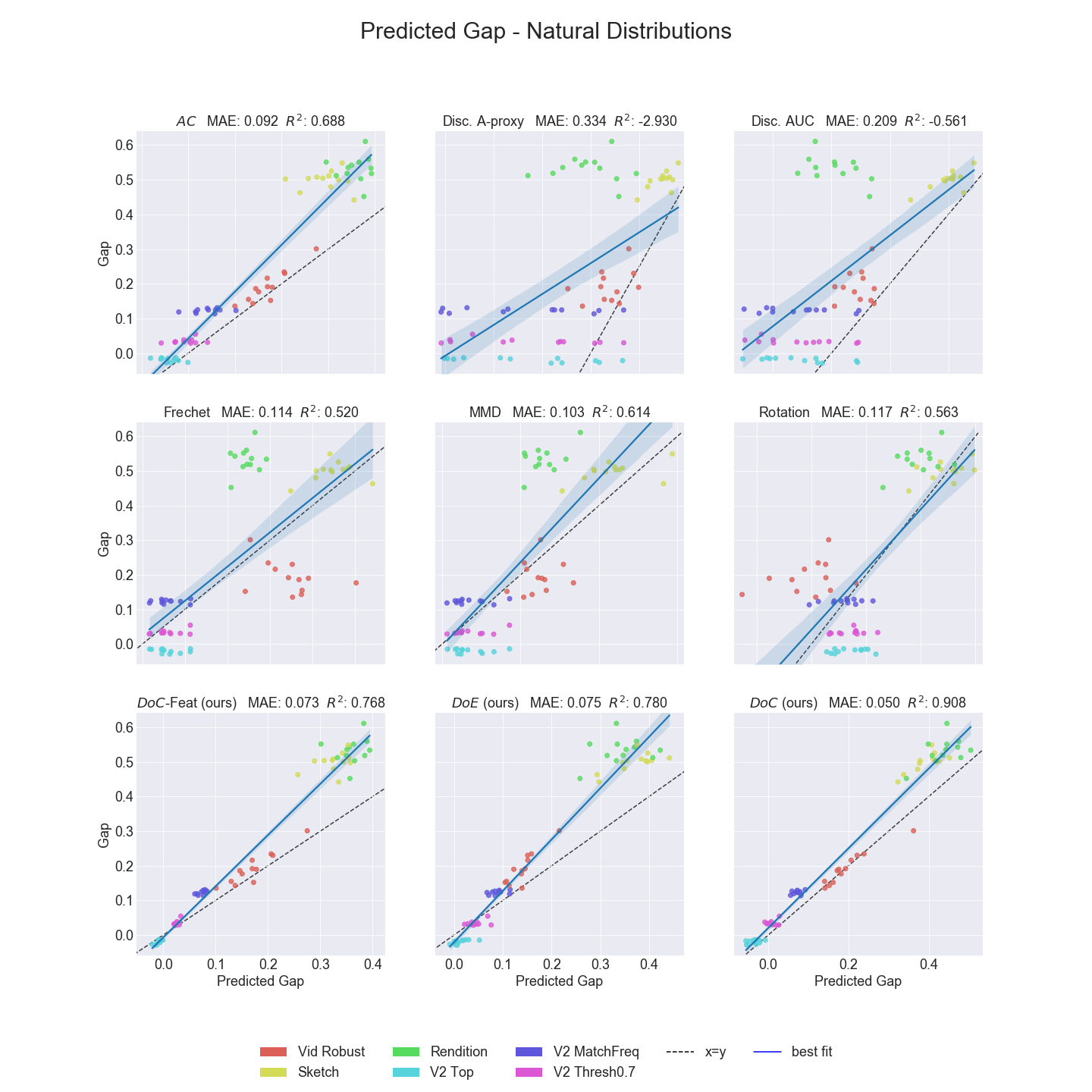}
\caption{
For each approach explored in this work, we plot the actual accuracy gap over the predicted accuracy gap across all twelve architectures.
We include MAE and $R^2$ score to give a since of aggregate fit.
Additionally, we plot the line of best fit in blue and the $x=y$ line which would represent perfect predictions as a dashed black line.
We see that even without any exposure to natural distribution shifts during calibration our approach $DoC$ does a good job of aligning predictions with the $x=y$ line.  
}
\label{fig:scatter-nat}
\end{figure}

Figure \ref{fig:scatter-all-raw}, shows the value of the featurization, $S$, of models across all test distributions ($Syn2$, $Natural$, $Adv$).
We note many of the same trends as in \ref{fig:scatter-nat-raw}, and the consistent clustering of ImageNet-A as an outlier group.
Based upon these visualizations, it appears that measures such as MMD and Frechet distance are non-informative, with regards to predicting accuracy under distribution shifts.
However, it is important to note that we learn model-specific regressors for predicting accuracy, and while these measures may not be informative for a global predictor, we see in Figure \ref{fig:scatter-nat} that when we observe these values for a single model (ResNet-18) , the correlations significantly increase.
We observe that the trends on the single model plot, mirror those observed in our regression experiments and highlight the ability of these approaches to learn an informative prediction model that works for various types of shift. 

\begin{figure}[ht]
\centering\includegraphics[width=0.8\textwidth]{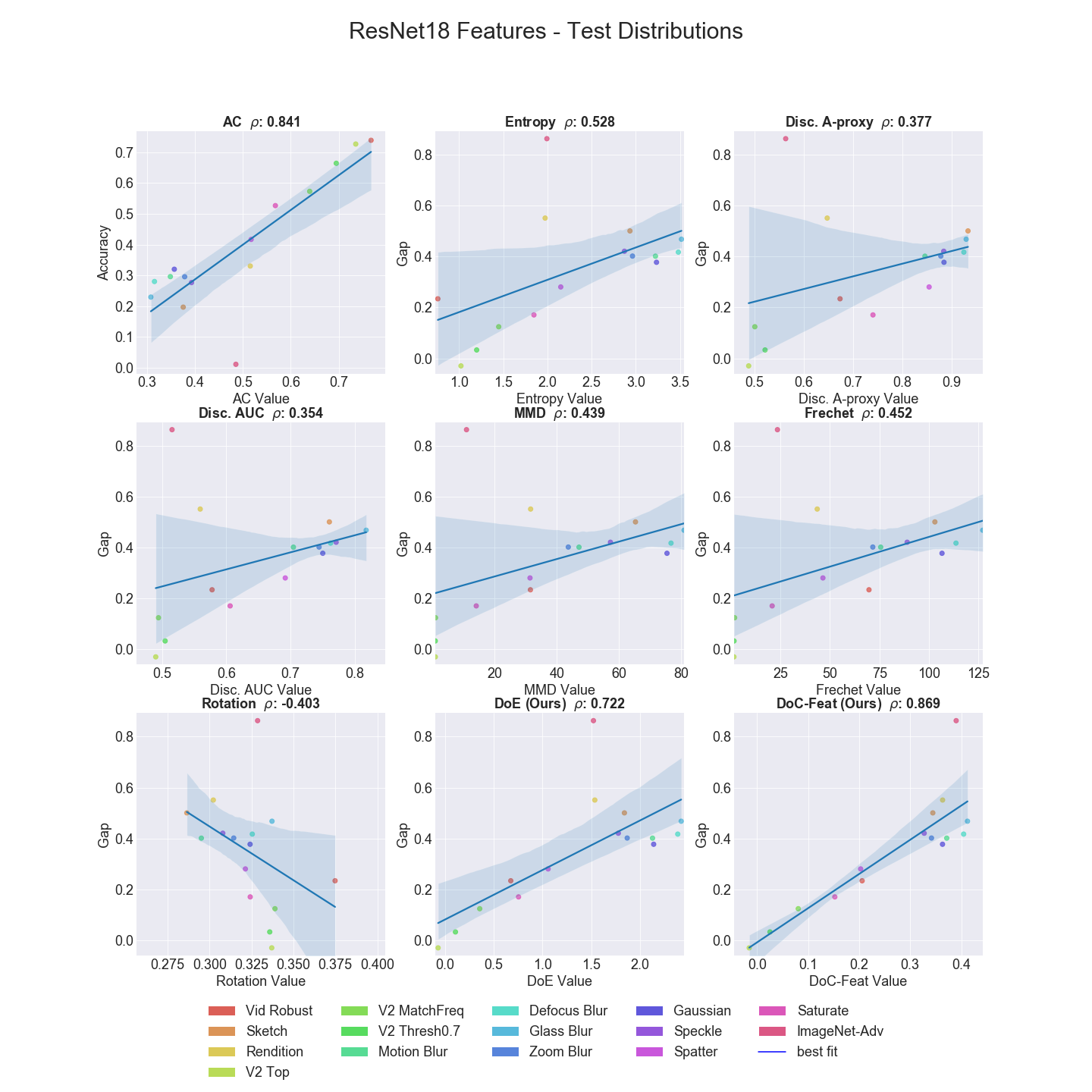}
\caption{
We plot the featurization value ($S$) and the corresponding accuracy gap for each approach explored in this work, as well as the line of best fit over all test set distribution on a ResNet-18 model.
We also include Pearson correlation coefficient ($\rho$), for each plot to capture a measure of how informative each feature could be with a perfect oracle. 
For $AC$, we plot the accuracy instead of the accuracy gap because it is a direct estimate.
}
\label{fig:scatter-res18-all-raw}
\end{figure}

\par

Figure \ref{fig:scatter-nat}, plots the predicted gaps and actual gaps over natural distribution shifts for each approach presented in this work and includes the line of best fit, MAE, and the coefficent of determination ($R^2$) to summarize how well each approach is doing.
Additionally, the line $x=y$ is included on the plot for reference to where the points would land if the predictions were perfect. 
We note that our confidence-based measures have the lowest MAE and highest $R^2$, and that predicted error tends to grow as the actual gaps increase.
For natural distribution shifts, we see that confidence-based measures tend to produce optimistic estimates with $y>x$ for nearly all points.
\par
\begin{figure}[ht]
\centering\includegraphics[width=0.8\textwidth]{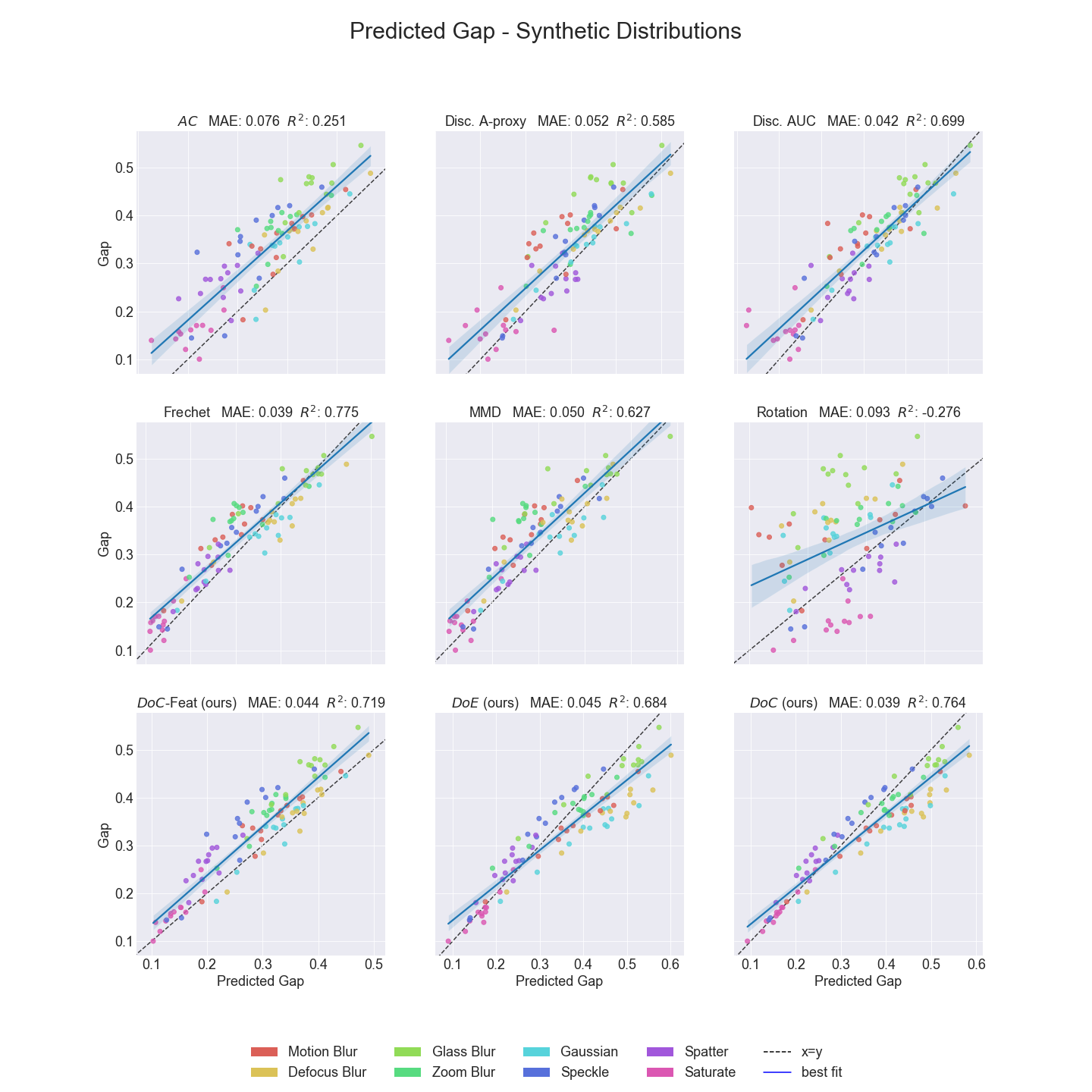}
\caption{
For each approach explored in this work, we plot the actual accuracy gap over the predicted accuracy gap across all twelve architectures over synthetic distributions $Syn2$.
We include MAE and $R^2$ score to give a since of aggregate fit.
Additionally, we plot the line of best fit in blue and the $x=y$ line which would represent perfect predictions as a dashed black line.
We see that even most approaches reliably predict accuracy over synthetic shifts when calibrated over a disjoint set of synthetic perturbations.
}
\label{fig:scatter-syn-reg}
\end{figure}

In Figure \ref{fig:scatter-syn-reg}, we visualize the predicted gaps of all approach over the held-out synthetic shifts, $Syn2$. 
It is important to note how well most all of the approaches fare at predicting accuracy on held-out synthetic shifts, with the exception of rotation prediction. 
The drastic difference in the performance of approaches on Figure \ref{fig:scatter-nat} and \ref{fig:scatter-syn-reg}, highlight a limitation of prior work studying distribution shift solely on synthetic or natural shifts and highlights the importance of understanding how these forms of shift relate to each other.
\par

\begin{figure}[ht]
\centering\includegraphics[width=0.8\textwidth]{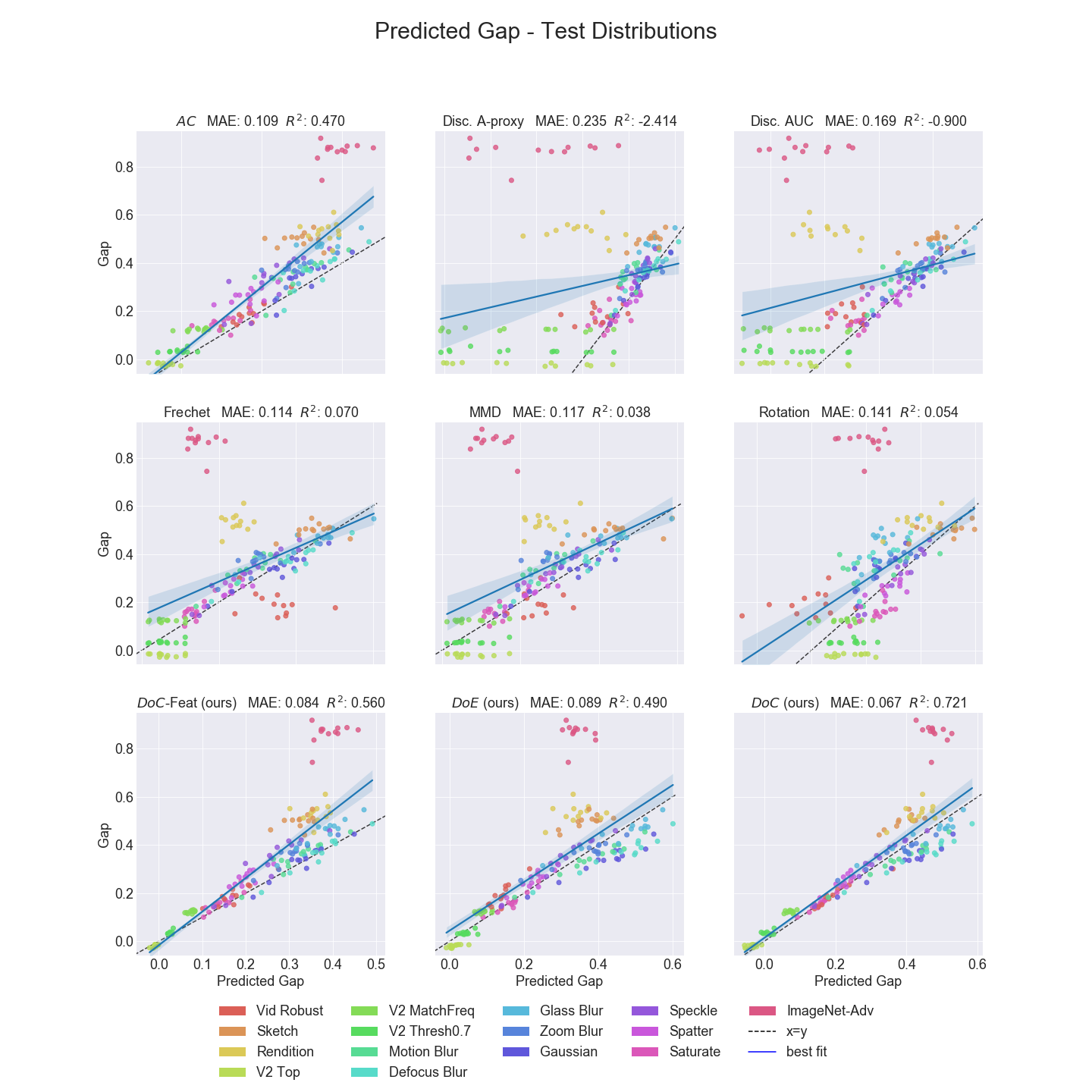}
\caption{
For each approach explored in this work, we plot the actual accuracy gap over the predicted accuracy gap across all twelve architectures over all distribution shifts in the test set ($Natural$, $Syn2$ $Adv.$).
We include MAE and $R^2$ score to give a since of aggregate fit.
Additionally, we plot the line of best fit in blue and the $x=y$ line which would represent perfect predictions as a dashed black line.
We note that while some approaches are able to capture both synthetic and natural distributions well, the adversarial examples in ImageNet-A continue to present a unique challenge.
}
\label{fig:scatter-all-reg}
\end{figure}

Figure \ref{fig:scatter-all-reg} show the predicted gaps and actual accuracy gaps over all held-out distribution shifts ($Syn2$, $Natural$, $Adv.$).
While confidence-based approaches still provide the best estimates of performance, it is important to note that despite having the lowest correlations in Figure \ref{fig:scatter-all-raw}, Frechet distance and MMD yield more accurate predictions on unseen shifts than Disc. AUC and Disc. A-proxy.
While some of the approaches capture natural and synthetic shifts near equally well, ImageNet-A instances universally present themselves as outliers for all approaches. 
In order to illustrate how well, the correlation present in a single model (ResNet-18) is captured based on our problem paradigm, Figure \ref{fig:scatter-nat} shows the results for the predicted gaps of this single model.
When we compare with the raw features in \ref{fig:scatter-res18-all-raw}, we see that our regression models do a relatively poor job of capturing the correlation present in discriminative distances, but do well in the other settings.

\begin{figure}[ht]
\centering\includegraphics[width=0.8\textwidth]{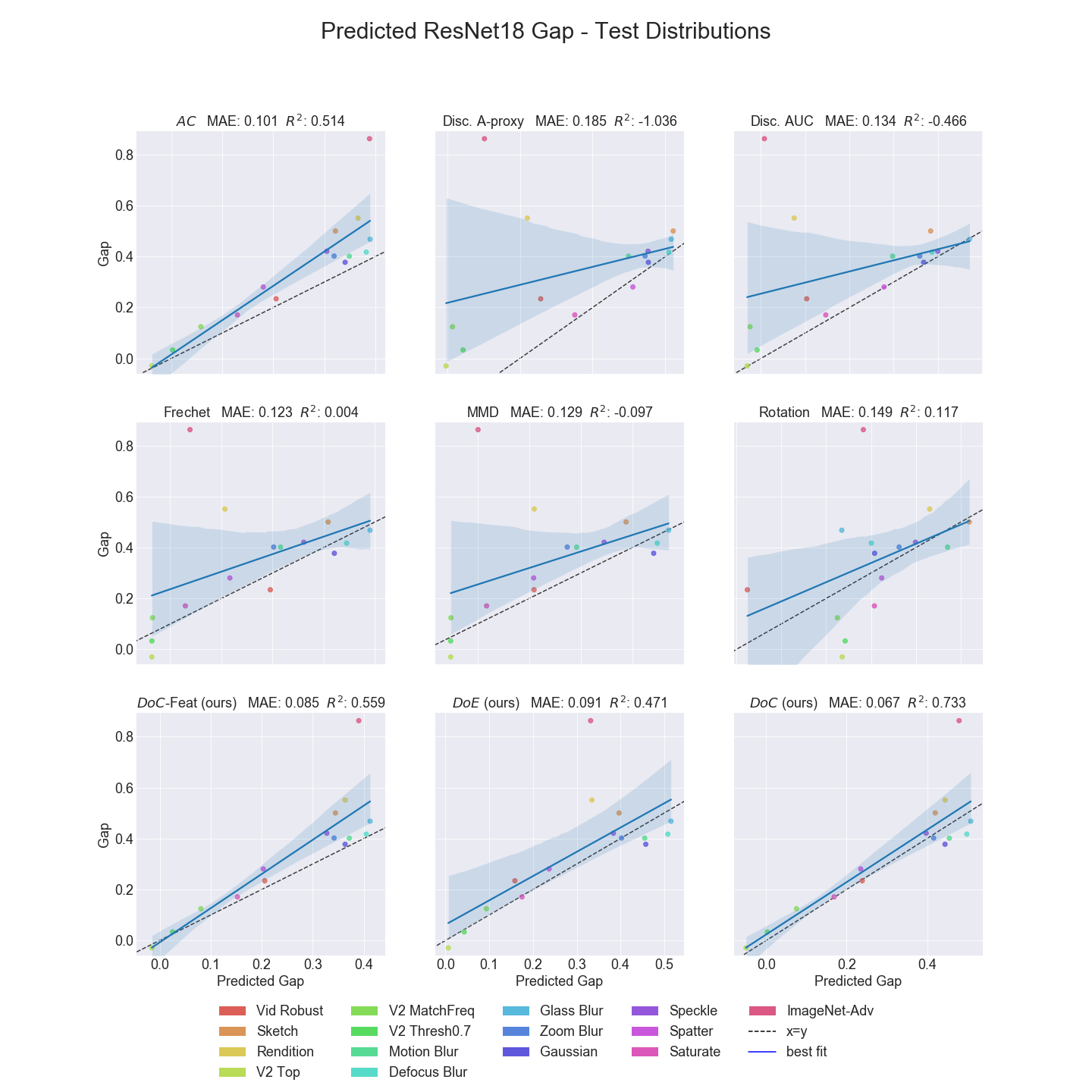}
\caption{
For each approach explored in this work, we plot the actual accuracy gap over the predicted accuracy gap on ResNet-18 over all distribution shifts in the test set ($Natural$, $Syn2$ $Adv.$).
We include MAE and $R^2$ score to give a since of aggregate fit.
Additionally, we plot the line of best fit in blue and the $x=y$ line which would represent perfect predictions as a dashed black line.
Comparing with \ref{fig:scatter-res18-all-raw}, we can see how well our models were able to fit the feature distributions.
}
\label{fig:scatter-res18-reg}
\end{figure}

\begin{table*}[ht]
\centering
\begin{tabular}{| c | c| *{4}{p{0.12\linewidth}|}} 
\hline
 &  & Syn1 &  Syn2 &  Natural & Adversarial \\  \hline
cal & Algorithm & MAE (std) & MAE (std) & MAE (std) &  MAE (std)  \\ \hline
\multirow{3}{*}{N/A} & Base Acc & 0.32 (0.069) & 0.328 (0.063) & 0.234 (0.087) &  0.865 (0.040)  \\ \cline{2-6}
 & $AC$ & 0.092 (0.042) & 0.075 (0.035) & 0.087 &  0.476 (0.047)  \\ \cline{2-6}
 & $DoC$-Feat & 0.058 (0.026) & 0.043 (0.017) & 0.072 (0.033) &  0.474 (0.043)  \\ \cline{1-6}
\multirow{5}{*}{Syn1} & Disc. A-proxy & - & 0.052 (0.017)  & 0.335 (0.19) & 1.109 (0.266) \\ \cline{2-6}
& Disc. AUC & - & 0.042 (0.016)  & 0.207 (0.094)  & 0.94 (0.10)  \\ \cline{2-6}
& Frechet & - & 0.0384 (0.013) & 0.113 (0.040)  & 0.713 (0.054)  \\ \cline{2-6}
& MMD & - & 0.049 (0.0178) &  0.104 (0.037) & 0.738 (0.066) \\ \cline{2-6}
& Rotation & - & 0.094 (0.039) & 0.115 (0.036) & 0.665 (0.075) \\ \cline{2-6}
& $DoE$ & -  & 0.045 (0.017)  & 0.075 (0.032)  & 0.524 (0.051) \\ \cline{2-6}
& $DoC$ & - & \textbf{0.039} (0.015) &  \textbf{0.050} (0.021) & 0.388 (0.055) \\ \cline{1-6}
\multirow{5}{*}{Syn2} & Disc. A-proxy & 0.070 (0.019)  & - & 0.261 (0.137)  & 1.04 (0.196)  \\ \cline{2-6}
& Disc. AUC & 0.056 (0.015) & -  & 0.133 (0.070) & 0.836 (0.098) \\ \cline{2-6}
& Frechet & 0.044 (0.013) & -  & 0.126 (0.038) & 0.661 (0.059)  \\ \cline{2-6}
& MMD & 0.052 (0.019) & -  & 0.111 (0.035) & 0.678 (0.072) \\ \cline{2-6}
& Rotation & 0.082 (0.032) & - & 0.189 (0.074) & 0.588 (0.089) \\ \cline{2-6}
& $DoE$ & 0.039 (0.016)  & -  & 0.096 (0.034)  & 0.537 (0.043)  \\ \cline{2-6}
& $DoC$ & \textbf{0.032}  (0.014)& -  & 0.066 (0.027) & 0.432 (0.040)  \\ \cline{1-6}
\multirow{5}{*}{Natural} & Disc. A-proxy & 0.163 (0.038)  & 0.134 (0.030)  & - & 0.734 (0.063)  \\ \cline{2-6}
& Disc. AUC & 0.145 (0.029) & 0.118 (0.030)  & - & 0.723 (0.052) \\ \cline{2-6}
& Frechet & 0.056 (0.025) & 0.045 (0.026)   & - & 0.691 (0.057)  \\ \cline{2-6}
& MMD & 0.083 (0.046) & 0.0596 (0.031) & - & 0.724 (0.056) \\ \cline{2-6}
& Rotation & 0.059 (0.022) & 0.098 (0.045) & - & 0.654 (0.052) \\ \cline{2-6}
& $DoE$ & 0.146 (0.068) & 0.186 (0.087)  & -  & 0.371 (0.065) \\ \cline{2-6}
& $DoC$ & 0.070 (0.025)  & 0.096 (0.036)  & - & \textbf{0.292} (0.049)  \\ \cline{1-6}

\end{tabular}
\caption{We show the mean absolute error (MAE) and standard deviation (std) in predicting accuracy over each approach given various calibration settings. We observe that $DoC$ produces the best overall prediction for each target distribution. $DoC$ also consistently produces the best predictions for natural distribution shifts regardless of calibration setting.}
\label{tab:main}
\end{table*}

\section{Baseline Algorithms}

In this section we elaborate on and present formulas used to compute the baseline algorithms discussed in Section \ref{sec:elsewhere}.

\paragraph{Maximum Mean Discrepancy.}
The distribution distance metric, Maximum Mean Discrepancy (MMD) \cite{borgwardt2006integrating} is commonly used in domain adaptation methods \cite{tzeng2014deep, long2017deep} and has even been shown to be correlated with target domain accuracy \cite{tzeng2014deep}. We use the following empirical estimate of MMD between base and target datasets.

\begin{equation}
  \text{MMD}(B^\prime, T^\prime) = \Big\| \frac{1}{|B^\prime|}\sum_{x \in B^\prime}{F^\prime(x)} - \frac{1}{|T^\prime|}\sum_{x \in T^\prime}{F^\prime(x)} \Big\|
\end{equation}

\paragraph{Rotation Prediction.}
Several self-supervised learning approaches \cite{gidaris2018unsupervised, henaff2020data} take advantage of rotation prediction as a useful auxiliary (pretext) task.
Recent work in adaptation \cite{sun2020test} and generalization \cite{reed2021selfaugment, hendrycks2019using} has shown rotation prediction to be an informative task in predicting generalization gaps and improving model performance on shifted domains.
As such we train linear models to predict which of the four rotations (0, 90, 180, 270) have been applied to an image, based on the model's featurization, $F^\prime$.
These models are trained over the base dataset $B$ and the accuracy and AUC are reported over the target datasets $T$.

\paragraph{Frechet Distance.}
Frechet distance is a popular method of comparing high-dimensional image distributions. 
Frechet Inception Distance \cite{heusel2017gans, zilly2019frechet} was popularized in the computer vision community as a method of evaluating the quality of generative models. Since Frechet Distance can be computed from summary statistics of the features extracted over the base and target domains, we compute it over the entirety of the datasets and ignore the train / validation / test data splits.
\begin{equation}
    \begin{aligned}
       \text{Frechet} = \|\bar{F^\prime}(B) - \bar{F^\prime}(T)\| +  tr\Big(\Sigma_{F^\prime(B)} + \\ \Sigma_{F^\prime(T)}  - 2(\Sigma_{F^\prime(B)}\Sigma_{F^\prime(T)})^{1/2} \Big)
    \end{aligned}
\end{equation}
where $\bar{F}'(X)$ denotes the average of the featurizations over the dataset $X$ and $ \Sigma $ represents the covariance matrix.

\paragraph{Discriminative distances.}
In theory and practice, a large number of domain adaptation  \cite{ganin2016domain,ben2006analysis, cortes2019adaptation, tzeng2017adversarial} reduce the error from domain shift by minimizing the ability to discriminate between the base distribution $\mathcal{B}$ and the target distributions $\mathcal{T}$, often dependent on a featurization $F^\prime$ of the classifier. 
In order to adequately estimate discriminative capacity, we create train, validation, and test splits on $B^\prime$ and $T^\prime$. 
We train linear classifiers to discriminate between the base dataset and target dataset based on train and validation splits.
Multilayer perceptrons (MLPs) are explored in the supplemental materials. 
The quality of this discriminator is evaluated over the separate test split to avoid overfitting. We capture several metrics from this discriminator such as accuracy, AUC, and A-proxy \cite{ben2006analysis} defined as:
\begin{equation}
   \text{A-proxy} = 2 \, (1- 2 \times \text{error}) \; .
\end{equation}

\paragraph{Temperature Scaling}
As we are are using the average confidence ($AC$)of uncalibrated models as our baseline, we also include results of these models calibrated over the base dataset, $B$, with Temperature Scaling \cite{guo2017calibration}. 
We use a single temperature parameter and optimize with negative log likelihood as in \cite{guo2017calibration}.

\section{Experimental Details}
To account for the different subsets of labels that our distribution shifts operate over, we compute the accuracy on ImageNet-Val for the subset of classes used by each shifted distribution and attempt to predict the change in accuracy with respect to this number. 
Because certain measures of distributional difference will be significantly impacted by the classes present, we compute our distributional difference, $S$, only over the instances that contain the classes present in our target distribution $T$. 
For some measures, we seek to optimize an auxilary model (discriminator) to estimate distribution distances. 
To prevent overfitting on these estimates, we train these models over $40\%$ of the base $B$ and target $T$ distributions. 
We tune the hyperparameters of these models based on an additional $10\%$ of these distributions, and report distance measures based on the remaining $50\%$ of the distributions. 
Table \ref{tab:summary_dist} presents methods used in this work as well as the specific data splits the measures were computed over and what values they are used to predict.

\begin{table}[ht]
    \centering
    \begin{tabular}{c|c|c}
         &  Eval Data & Estimated Value \\ \hline
        Base Acc &  $B^\prime$ & Target Acc ($A_T^B$) \\
        AC &  $T^\prime$ & Target Acc ($A_T^B$) \\
        AC TempScaling & $T^\prime$ & Target Acc ($A_T^B$) \\
        Frechet &  $B^\prime, T^\prime$ & $\Delta{\mathbf{Acc}}(B,T)$   \\
        Disc. A-Proxy &  $B^\prime_{\textit{test}}, T^\prime_{\textit{test}}$ & $\Delta{\mathbf{Acc}}(B,T)$ \\
        Disc. AUC &  $B^\prime_{\textit{test}}, T^\prime_{\textit{test}}$ & $\Delta{\mathbf{Acc}}(B,T)$  \\
        MMD &  $B^\prime, T^\prime$ & $\Delta{\mathbf{Acc}}(B,T)$  \\
        Rotation  & $T^\prime$ & $\Delta{\mathbf{Acc}}(B,T)$  \\
        $DoE$  & $B^\prime, T^\prime$ & $\Delta{\mathbf{Acc}}(B,T)$  \\
        $DoC$  & $B^\prime, T^\prime$ & $\Delta{\mathbf{Acc}}(B,T)$  \\
        $DoC$-Feat  & $B^\prime, T^\prime$ & $\Delta{\mathbf{Acc}}(B,T)$  \\
    \end{tabular}
    \caption{Summary table showing all methods of accuracy prediction used in this work along with what data is used to compute measures of difference and what accuracy value they are used to predict.}
    \label{tab:summary_dist}
\end{table}

\section{Res-Ensemble}
Deep Ensembles were identified as the calibration approach with the best performance under distribution shift in \cite{ovadia2019can}. 
As such we used an ensemble of Resnet Architectures to serve as a calibration baseline. Our Res-Ensemble model is comprised of pretrained Resnet-18, Resnet-34, Resnet-50, Resnet-101, and Resnet-152 architectures.

